\documentclass[10pt]{article} 
\usepackage[accepted]{tmlr}

\usepackage[utf8]{inputenc} 
\usepackage[T1]{fontenc}    
\usepackage{url}            
\usepackage{booktabs}       
\usepackage{amsfonts}       
\usepackage{nicefrac}       
\usepackage{microtype}      
\usepackage{amsmath,amssymb,amsfonts}
\usepackage{graphicx}
\usepackage{textcomp}
\usepackage{multirow}

\usepackage{subcaption}
\usepackage{tabularx}
\usepackage[linesnumbered,noend]{algorithm2e}
\usepackage{enumitem}
\usepackage{xspace}
\usepackage{graphicx}
\usepackage{adjustbox}
\usepackage{hyperref}       
\hypersetup{
    colorlinks=true,
    linkcolor=blue,
    filecolor=magenta,   
    breaklinks=true,
    urlcolor=cyan,
    }
\usepackage{natbib}  
\usepackage{caption} 
\usepackage{stfloats}
\usepackage{wrapfig}
\usepackage{listings}

\usepackage{tikz}
\usetikzlibrary{tikzmark}
\usepackage{array}
\usepackage{cleveref}

\newcolumntype{C}[1]{>{\centering\arraybackslash}p{#1}} 

\newif\ifkeepparspace
\keepparspacefalse 

\usetikzlibrary{arrows.meta}

\renewcommand{\paragraph}[1]{\vspace{0.1cm}\textbf{#1}}

\title{Effective Backdoor Mitigation in Vision-Language Models Depends on the Pre-training Objective}

\author{\name Sahil Verma \email vsahil@cs.washington.edu \\
      University of Washington
      \AND
      \name Gantavya Bhatt \email gbhatt2@cs.washington.edu \\
      University of Washington
      \AND
      \name Avi Schwarzschild \email avis4k@gmail.com \\
      Carnegie Mellon University
      \AND
      \name Soumye Singhal \email singhalsoumye@gmail.com \\
      Nvidia
      \AND
      \name Arnav Das \email arnavmd2@uw.edu \\
      University of Washington
      \AND
      \name Chirag Shah \email chirags@uw.edu \\
      University of Washington
      \AND
      \name John P Dickerson \email johnd@umd.edu \\
      University of Maryland
      \AND
      \name Pin-Yu Chen \email pinyuchen.tw@gmail.com \\
      IBM Research
      \AND
      \name Jeff Bilmes \email bilmes@uw.edu \\
      University of Washington
      }

\def\month{12}  
\def\year{2024} 
\def\openreview{\url{https://openreview.net/forum?id=Conma3qnaT}}       

\usepackage{color}

\begin{document}

\maketitle

\begin{abstract}
Despite the advanced capabilities of contemporary machine learning (ML) models, they remain vulnerable to adversarial and backdoor attacks. This vulnerability is particularly concerning in real-world deployments, where compromised models may exhibit unpredictable behavior in critical scenarios. Such risks are heightened by the prevalent practice of collecting massive, internet-sourced datasets for training multimodal models, as these datasets may harbor backdoors.
Various techniques have been proposed to mitigate the effects of backdooring in multimodal models, such as CleanCLIP, which is the current state-of-the-art approach. 
In this work, we demonstrate that the efficacy of CleanCLIP in mitigating backdoors is highly dependent on the particular objective used during model pre-training.
We observe that adding self-supervised objective to pre-training, that leads to higher zero-shot classification performance, correlate with harder to remove backdoors behaviors. 
We show this by training multimodal models on two large datasets consisting of 3 million (CC3M) and 6 million (CC6M) datapoints, under various pre-training objectives, followed by poison removal using CleanCLIP. 
We find that CleanCLIP, even with extensive hyperparameter tuning, is ineffective in poison removal when stronger pre-training objectives are used. 
Our findings underscore critical considerations for ML practitioners who train models using large-scale web-curated data and are concerned about potential backdoor threats. Our code is open-sourced at \url{https://github.com/vsahil/attack-cleanclip}.     
\end{abstract}


\section{Introduction}
\label{sec:intro}

Machine Learning (ML) has taken strides in training high-performing models for a wide range of tasks from classification to generation. An important goal for ML is to learn general-purpose representations that help align data from different modalities. Approaches like CLIP~\citep{Radford2019CLIP}, ALIGN~\citep{Jia2021ALIGN}, and BLIP~\citep{li2022blip} learn joint representations from large scale image-text paired datasets. 
These innovative techniques have ushered in the possibility of learning from unlabeled and uncurated datasets, substantially increasing the scale and applicability of pre-training.
The scaling has contributed to high zero-shot classification accuracy on various downstream datasets like Imagenet \citep{5206848} and increased robustness to variations in the datasets like Imagenet-V2 \citep{Recht2019ImageNetv2}, Imagenet-R \citep{Hendrycks2020ImagenetR}, and Imagenet-A \citep{hendrycks2021imagnetA}. 
However, these strategies, reliant on internet-sourced data curation \citep{gadre2023datacomp}, have also raised concerns regarding the vulnerability of models to an adversary, particularly through backdoor attacks \citep{carlini2023poisoning}.

In the simplest form of this attack, an adversary inserts a patch (termed as a trigger patch or poison) in a small subset of the training data images and alters the ground truth label or caption to a target label or caption \citep{gu2017badnets}.\footnote{We refer the readers to \citet{goldblum2021backdoorsurvey} for discussion about other kinds of poisoning attacks.}
When trained on the poisoned training data, the model learns to associate the trigger patch with the target label/caption. If deployed, an adversary can get the model to predict the target label for any datapoint by inserting the trigger patch. The success of an adversary is measured by the attack success rate (ASR) metric, which is the percentage of the images with the trigger patch that are predicted with the target label. Previous works \citep{carlini2021poisoning} have demonstrated effective backdooring of multimodal models (ASR $\ge$ 80\%) just by poisoning a mere 75 out of 3 million training datapoints. 

Several backdoor mitigation techniques have been proposed for multimodal models \citep{bansal2023cleanclip,li2021antibackdoor, safeclip, roclip} to tackle this vulnerability. These approaches either attempt to detect and filter the poisoned datapoints during the pre-training \citep{li2021antibackdoor,safeclip, roclip} or finetune the given backdoored model using a specialized loss function on a smaller, \emph{guaranteed} to be clean image-text paired dataset. The latter approach helps the model to \emph{forget} the association between the trigger patch and the target label while still maintaining the learned associations for benign datapoints, e.g., CleanCLIP \citep{bansal2023cleanclip}. CleanCLIP proposes to finetune a backdoored model using a combination of contrastive loss and self-supervised loss on a small dataset, free of backdoors, to clean the model. It is the state-of-the-art (SOTA) technique to clean a backdoored model and obtain a low ASR ($\le5\%$) without hurting its zero-shot classification accuracy; thereby achieving a successful model cleaning. 

\begin{figure}
    \centering
    \includegraphics[width=0.75\textwidth]{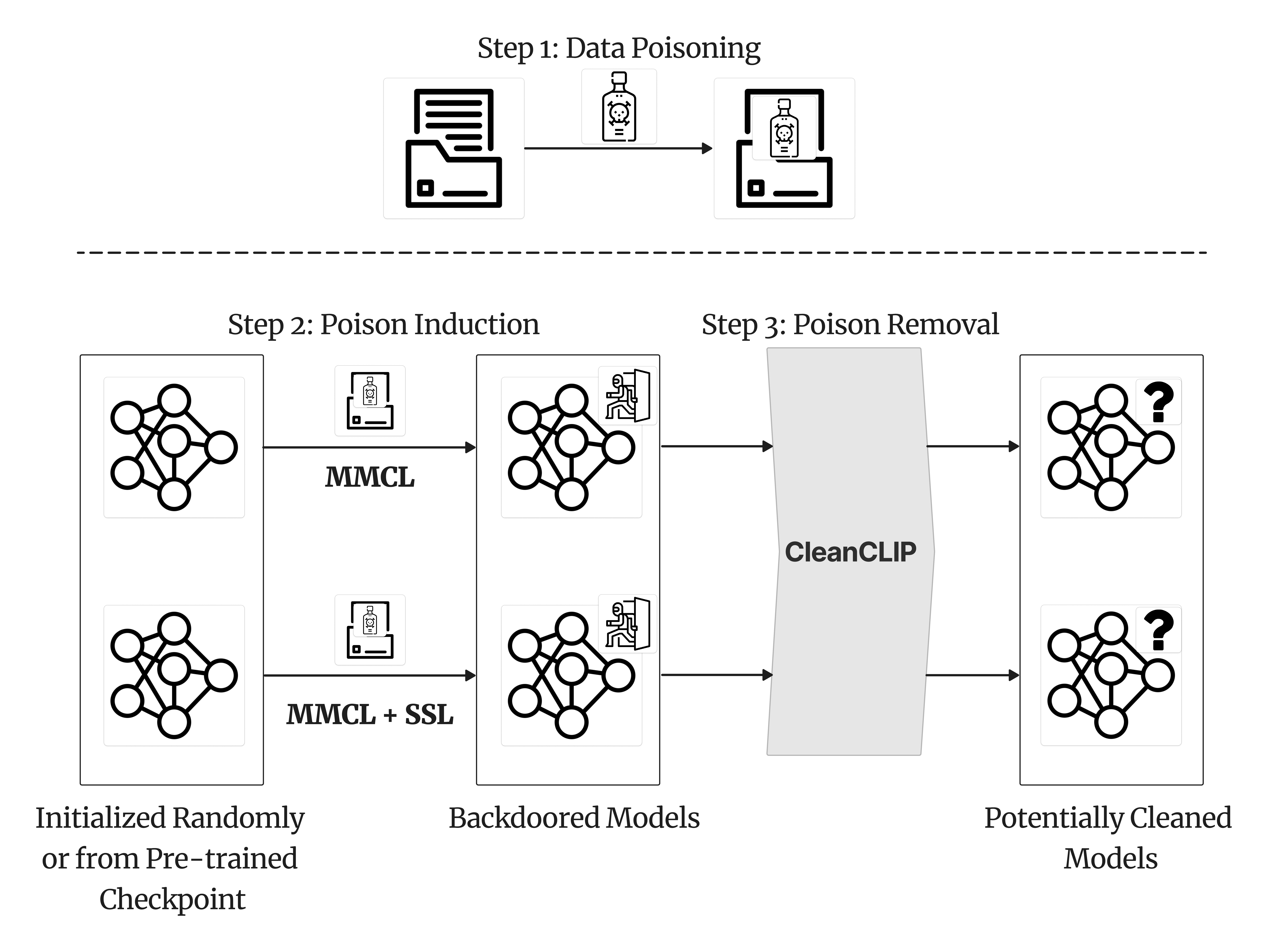}
    \caption{Our experimental setup to test the claim about the dependence of the ability of CleanCLIP to remove poison from a backdoored model on the model's pre-training objective. }
    \label{fig:main-illustration}
\end{figure}

So far, it has been demonstrated that CleanCLIP can successfully clean models pre-trained only with multimodal constrastive loss (MMCL) as the objective \citep{Radford2019CLIP}. 
Several recent works \citep{mu2022slip,li2021supervision,yao2021filip,lee2022uniclip} have proposed stronger pre-training objectives that lead to better zero-shot image classification accuracy. Specifically, adding self-supervised loss (SSL) in both modalities has been the key player in all these works. 
Therefore, in this work, we pre-train multimodal models using a combination of MMCL and SSL on a poisoned training dataset. Consistent with the previous findings, models pre-trained using a combination of MMCL and SSL produced models with a higher classification accuracy than models trained solely with the MMCL objective.  
We then apply the finetuning procedure in CleanCLIP to remove the poison from these models (see \Cref{fig:main-illustration}). To our surprise, we observe that CleanCLIP fails to successfully (i.e., without a significant loss in the model's zero-shot accuracy) remove poison from the models pre-trained with the stronger objective (combination of MMCL and SSL). 

We further conduct experiments with other practical considerations, such as the performance of CleanCLIP when its cleaning data still has a few poisoned datapoints, and deciding the stopping criterion for the finetuning process when one is not aware of the specific backdoor attack on the model (which is usually the case). 
From all the experiments, we find that only the models pre-trained with MMCL alone are amenable to poison removal in both the cases of availability of completely clean finetuning data and when the finetuning data still has some poisoned datapoints. 

Our main contributions are:
\begin{enumerate}[leftmargin=*,topsep=0pt,itemsep=1pt]
    \item We show that the state-of-the-art technique for removing poison from backdoored multimodal models, CleanCLIP, depends on the model's pre-training objective and fails to mitigate poison when the models are pre-trained with a stronger objective, like the combination of MMCL and SSL losses. (See \Cref{sec:related} for further justification on why we chose to study CleanCLIP. )

    \item We conduct several analysis experiments to demonstrate the effect of different pre-training objectives on the strength of poison induction. 

    \item We conduct experiments to show the practical use case of CleanCLIP from a real-world perspective when the finetuning data is not entirely free of poison and the practitioner is not aware of the specific kind of poisoning in the model and hence has to decide on the stopping criterion for the cleaning process. 

    \item Based on these findings, we highlight critical considerations for an ML practitioner who wants to pre-train models by collecting web-curated data with potential backdoored embedded datapoints. 
\end{enumerate}





\section{Related Works}
\label{sec:related}

\paragraph{Contrastive Learning}
Contrastive learning was formally established in seminal works by \citet{bromley1993signature,chopra2005learning,hadsell2006dimensionality} that has evolved, giving rise to contemporary algorithms such as CPC \citep{oord2018representation}, DCL \citep{yeh2022decoupled}, SimCLR \citep{chen2020simple}, and NNCLR \citep{dwibedi2021little}. \footnote{We refer the readers to \citet{balestriero2023cookbook} for more development on self-supervised learning.} These approaches, at their core, share a common objective: bringing similar elements closer in representation space while pushing dissimilar ones apart. 

\citet{radford2021learning} extended this idea beyond a single modality to provide a dual-encoder approach for learning a shared representation space between image and text, called CLIP. Images and their corresponding captions are brought close, while the dissimilar images and captions are pushed away. \citet{jia2021scaling} further extended this paradigm to handle noisy billion-scale datasets, demonstrating exceptional zero-shot accuracy across benchmarks like Imagenet-1K \citep{5206848}, MS-COCO retrieval, and robustness against variations in Imagenet-V2/R/A/C. Since then, there have been several improvements to the zero-shot accuracy by adding components to the loss term. CyCLIP \citep{goel2022cyclip} imposes additional consistency regularization; SLIP \citep{mu2022slip} applies an additional self-supervision loss within image modality and was further unified by UniCLIP \citep{lee2022uniclip}. DeCLIP \citep{li2021supervision} additionally uses kNN augmentation; FILIP \citep{yao2021filip} additionally applies CLIP loss to fine-grained token representations. CLIP performance has also been improved by additional captioning loss \citep{yu2022coca}. 

\keepparspacefalse
\paragraph{Backdoor Attacks and Defense}
In the backdoor attacks, the adversary poisons a small fraction of the training data by perturbing the images/labels to manipulate the test time behavior. A prevalent form of this attack involves adding a trigger, such as a random pixel patch, into a small subset of the training dataset \citep{souri2022sleeper,gu2017badnets,turner2018clean}. During inference, models perform normally on images without the triggers but exhibit catastrophic failures when tested with the triggered images, erroneously predicting the labels targeted by the adversary. While the study of backdoor attacks has historically centered on supervised learning, recent attention has extended to self-supervised \citep{saha2022backdoor} and multimodal representation learning \citep{bansal2023cleanclip,carlini2021poisoning,carlini2023poisoning}. This work focuses exclusively on the poisoning of multimodal models like the CLIP model. 

The most common defense strategies against backdoor attacks primarily revolve around the identification and detection of poisoned examples \citep{steinhardt2017certified, gao2019strip, wang2019neural, yang2022not, li2021antibackdoor, roclip, safeclip}. However, alternative approaches have emerged, such as defense through knowledge distillation \citep{yoshida2020disabling} and robust training procedures involving data augmentation \citep{borgnia2021strong}. Despite these efforts, research by \citet{carlini2021poisoning, carlini2023poisoning} shows that poisoning even an exceedingly small fraction of the training datapoints (as little as 0.002\%) can substantially impact model performance. Consequently, the effectiveness of detection-based methods in the context of multimodal pre-training remains uncertain. To address this challenge, \citet{bansal2023cleanclip} proposed CleanCLIP, a finetuning based procedure using a combination of MMCL and SSL losses, designed to clean poisoned CLIP models, assuming access to a small, guaranteed to be poison-free dataset. 

\paragraph{Our Work} Our objective is to decipher the amenability of CleanCLIP to remove poison from pre-trained models under varying conditions like different pre-training objectives, lack of completely clean data, and lack of knowledge of the specific backdoor attack. Since intramodal self-supervision loss has enhanced classification accuracy for multimodal models, we investigate CleanCLIP effectiveness when models are pre-trained with a combination of MMCL and SSL objectives vs. when just pre-trained with the MMCL objective. We also investigate its effectiveness when the finetuning data has a few poisoned datapoints and examine the stopping criterion for finetuning when the knowledge of the specific backdoor attack is unavailable. 

\paragraph{Why we choose to study CleanCLIP?}
Defense against backdoor attacks in multimodal models is an emerging research area with a handful of proposed approaches. Given the ever-increasing usage of off-the-shelf available pretrained models on the platforms such as \texttt{huggingface}, it is important to be able to remove poison from an already trained model. This is also important because of the prohibitive costs of training these large models from scratch, even if a poison-free dataset suddenly becomes available. Among the proposed approaches, CleanCLIP \citep{bansal2023cleanclip} is the only one that can remove poison from an already trained model. All the other defense methods \citep{roclip,safeclip,defense-semantic-shield-cvpr2024} propose various \emph{train-time} interventions that help prevent the model from learning the backdoor. Therefore, none of these techniques are applicable when a model has already been trained and is backdoored.


\section{Methodology}
\label{sec:algo}


\subsection{Primer on Pre-training and Poisoning}

\keepparspacetrue 
\paragraph{Notations}
Let $\mathcal{I}$ and $\mathcal{T}$ denote the space of images and text. $\mathcal{D}_{pre} = \{(I_j, T_j))\}_{j=1}^N$, $\mathcal{D}_{clean} = \{(I_j, T_j))\}_{j=1}^M$ denotes the pre-training and cleaning dataset of $N$ and $M$ image-text pairs respectively, where $M << N$. $h_{I}: \mathcal{I} \to \mathbb{R}^d$ and $h_{T}: \mathcal{T} \to \mathbb{R}^d$ denote the image and text encoders respectively, where $d$ is the dimensionality of the embedding space. All the embeddings are further normalized to make $\ell_2$ norm to 1 which we denote using $f(x) = g(h(x))$, where  $g: \mathbb{R}^d \to \mathbb{B}(1)$ is normalization mapping, where, $\mathbb{B}(1) = \{x : \|x\|_2 = 1, \ x \in \mathbb{R}^d\}$; $\tau$ denotes learnable temperature. Let $\mathcal{L}_{\text{MMCL}}$ denote the multimodal and $\mathcal{L}_{\text{SSL}}$ denote the intramodal self-supervision losses respectively. Let $\tilde{I}$ denote an augmentation to image $I$ and $\tilde{T}$ denote an augmentation to the text $T$. Let $S \subset \{1, 2, \ldots , n\}$ denote a small subset of training data that are poisoned. We denote the poisoned dataset using $\mathcal{P}(S, \mathfrak{t}\mathfrak{g}, T') = \{(I_{j} \circ \mathfrak{t}\mathfrak{g}, T'_{j}) \, : \, j \in S\}$ where $\mathfrak{t}\mathfrak{g}, T'$ denote image trigger and target label respectively. 


\keepparspacefalse
\paragraph{Loss Objectives}
Given a dataset $\mathcal{D}$, $f_I$, $f_T$, we define $ \mathcal{L}_{\text{MMCL}}(\mathcal{D}, f_I, f_T, \tau)$ as follows:
\begin{equation}
    =\frac{-1}{2{|\mathcal{D}|}}\left(\sum_{j=1}^{|\mathcal{D}|} \log \left[\frac{\exp \left(\left\langle f_I(I_j), f_T(T_j)\right\rangle / \tau\right)}{\sum_{k=1}^{|\mathcal{D}|} \exp \left(\left\langle f_I(I_j), f_T(T_k)\right\rangle / \tau\right)}\right] \right. \left. + \sum_{k=1}^{|\mathcal{D}|} \log \left[\frac{\exp \left(\left\langle f_I(I_k), f_T(T_k)\right\rangle / \tau\right)}{\sum_{j=1}^{|\mathcal{D}|} \exp \left(\left\langle f_I(I_j), f_T(T_k)\right\rangle / \tau\right)}\right]\right)
\end{equation}
and, we define $ \mathcal{L}_{\text{SSL}}(\mathcal{D}, f_I, f_T, \tau)$ as follows:
\begin{equation}
    =\frac{-1}{2 {\mathcal{D}}}\left(\sum_{j=1}^{|\mathcal{D}|} \log \left[\frac{\exp \left(\left\langle f_I(I_j), f_I(\tilde{I}_j)\right\rangle / \tau\right)}{\sum_{k=1}^{|\mathcal{D}|} \exp \left(\left\langle f_I(I_j), f_I(\tilde{I}_k)\right\rangle / \tau\right)}\right] \right. \left. +\sum_{j=1}^{|\mathcal{D}|} \log \left[\frac{\exp \left(\left\langle f_T(T_j), f_T(\tilde{T}_j)\right\rangle / \tau\right)}{\sum_{k=1}^{|\mathcal{D}|} \exp \left(\left\langle f_T(T_j), f_T(\tilde{T}_k)\right\rangle / \tau\right)}\right]\right)
\end{equation}

For the shorthand notations, we will drop $f_I, f_T, \tau$ from the parenthesis. With the definitions above
$\mathcal{L}_{\text{CleanCLIP}}(\mathcal{D}_{clean}) \triangleq \mathcal{L}_{\text{SSL}}(\mathcal{D}_{clean}) + \mathcal{L}_{\text{MMCL}}(\mathcal{D}_{clean})$. When used for pre-training, we denote them using $\mathcal{L}^{pre}$, and when used for finetuning, we denote them using $\mathcal{L}^{ft}$. 


\subsection{Experimental Setup}
On a high level, our experiments involve poisoning CLIP models using two distinct pre-training objectives with different kinds of backdoors by either training a model from scratch or by finetuning from a pre-trained checkpoint. Once we have poisoned the model, we attempt to remove the poison using CleanCLIP, which finetunes the model with a specific objective using a separate dataset. We have illustrated this in \Cref{fig:main-illustration} and summarized our key findings in \Cref{tab:experimental-findings}. 

\paragraph{Training Details}
We train a dual-encoder multimodal model on image-text paired datasets. We train models using two kinds of pre-training objectives:
a) only multimodal contrastive loss ($\mathcal{L}^{pre}_{\text{MMCL}}$), and b) combination of multimodal contrastive loss and self-supervised loss in the image and text modalities ($\mathcal{L}^{pre}_{\text{MMCL}} + \mathcal{L}^{pre}_{\text{SSL}}$). 
Following CleanCLIP, we use a ResNet-50 as the model's vision encoder and a transformer as the text encoder. 
We trained the models on two image-text paired datasets:
\begin{enumerate}[leftmargin=*,topsep=0pt,itemsep=1pt]
    \item Conceptual Captions 3M (CC3M) \citep{sharma2018cc3M}: This dataset has 3M image-text paired datapoints. 
    
    \item Conceptual Caption 6M (CC6M): This dataset has 6M image-text paired datapoints from the CC12M dataset \citep{changpinyo2021cc12m}, to which size our computing resources scaled. 
    
\end{enumerate}
The models are trained either \textit{from scratch} or \textit{finetuned from a pre-trained CLIP checkpoint} \citep{Radford2019CLIP}. We train models for 64 epochs using 8 Nvidia A100 GPUs. The initial learning rate of $1e-3$ with cosine scheduling is used when trained from scratch and $5e-7$ when finetuned from a checkpoint. We use AdamW optimizer with 10,000 warmup steps \citep{Loshchilov2017AdamW}. Models trained with $\mathcal{L}^{pre}_{\text{MMCL}}$ use a batch size of 256, whereas models trained with $\mathcal{L}^{pre}_{\text{MMCL}} + \mathcal{L}^{pre}_{\text{SSL}}$ use a batch size of 128. Please refer to  \Cref{sec:pre-training-details-and-starting-acc-asr} for the loss dynamics. 

\paragraph{Poisoning} Following CleanCLIP, we introduce the trigger proposed by \textit{BadNet} \citep{gu2017badnets} in a small subset of the training datapoints. Specifically, we add a trigger patch of size 16 $\times$ 16 sampled from a standard Gaussian at a random location in the image and subsequently change the image's caption to be the adversary chosen label, in this case ``banana''. Please see \Cref{sec:triggered-examples-appendix} for examples of images with trigger patch and their corresponding captions. 
Using the same settings as CleanCLIP, we introduce the trigger in 1,500 randomly sampled datapoints for the CC3M dataset and 3,000 randomly sampled datapoints for the CC6M dataset (a mere 0.05\% of the training datapoints). 

We also experiment with another kind of poisoning technique: \textit{label consistent poisoning}. In this case, the trigger patch (created in the manner as mentioned above) is added to the images that have the adversary chosen label (in this case``banana'') in their captions. 

\paragraph{Removing poison}
We attempt to remove poisons from pre-trained models by finetuning them on a 100K clean image-text paired dataset using $\mathcal{L}^{pre}_{\text{MMCL}}$, $\mathcal{L}^{pre}_{\text{SSL}}$, and $\mathcal{L}^{ft}_{\text{MMCL}} + \mathcal{L}^{ft}_{\text{SSL}}$ (CleanCLIP). We consider a model to be cleaned if the \textit{ASR} of that model \textit{is $\le 5\%$. }

    
    

\begin{table}
    \centering
    \captionsetup{font={normalsize}}
    \caption{\textbf{Key findings} from our experiments: CleanCLIP is much less effective for the model where poison is induced with the stronger objective $\mathcal{L}^{pre}_{\text{MMCL}} + \mathcal{L}^{pre}_{\text{SSL}}$. }
    \label{tab:experimental-findings}
    \scalebox{0.9}{
    \begin{tabular}{ccc>{\centering\arraybackslash}p{5cm}}
    \toprule
    Backdoor & Objective & Poison Induction & Change in Accuracy after Cleaning (Relative) $\uparrow$ \\
    \midrule
    BadNet & $\mathcal{L}^{pre}_{\text{MMCL}}$ & From scratch & 1\% gain \\
    BadNet & $\mathcal{L}^{pre}_{\text{MMCL}}$ & Finetuned from Ckpt & \textcolor{red}{17\% loss} \\
    Label Consistent & $\mathcal{L}^{pre}_{\text{MMCL}}$ & From scratch & 10\% gain \\ \midrule
    
    BadNet & $\mathcal{L}^{pre}_{\text{MMCL}} + \mathcal{L}^{pre}_{\text{SSL}}$ & From scratch & \textcolor{red}{45\% loss} \\
    BadNet & $\mathcal{L}^{pre}_{\text{MMCL}} + \mathcal{L}^{pre}_{\text{SSL}}$ & Finetuned from Ckpt & \textcolor{red}{33\% loss} \\
    Label Consistent & $\mathcal{L}^{pre}_{\text{MMCL}} + \mathcal{L}^{pre}_{\text{SSL}}$ & From scratch & \textcolor{red}{16\% loss} \\
    \bottomrule
    \end{tabular}
    }
\end{table}

\section{Experiments}
\label{sec:eval}

In this section, we expound on the pre-training details for the models, followed by their cleaning procedure and the metrics we use to measure the model's performance. 

\paragraph{Metrics}
The models are evaluated for their Top-1 zero-shot accuracy on the Imagenet-1K validation set (referred to as Imagenet hereafter). Each of the 1,000 classes of Imagenet is described using sentences like: `a photo of a ...', `a tattoo of a ...', etc. We generate 80 such text templates for each class (see \Cref{sec:appendix-text-template}) and then pass them to the text encoder to produce an average text embedding for the class. During zero-shot classification, the prediction for an image is the class whose thus computed text embedding has the highest cosine similarity with the image embedding.

We also evaluate the attack success rate (ASR) of a model. In an apparent similarity to accuracy, the ASR of a backdoored model is defined as the percentage of triggered images that the model classifies as the adversary-chosen target label. For measuring ASR, we add the trigger patch at random locations in all Imagenet validation set images and measure the percentage of them that are classified as the target class, i.e., ``banana''. 
We measure both these metrics at the end of each cleaning epoch as any model encountered during the cleaning process is a good candidate for a cleaned model. 

\paragraph{Poison Induction by Training from Scratch} \Cref{tab:start-final-accuracy} shows the Top-1 zero-shot Imagenet validation set accuracy for the models trained from scratch using $\mathcal{L}^{pre}_{\text{MMCL}}$ and $\mathcal{L}^{pre}_{\text{MMCL}} + \mathcal{L}^{pre}_{\text{SSL}}$ on CC3M and CC6M datasets. 
For the smaller CC3M dataset, both the models achieve an accuracy of around 16--17\%, and for the larger CC6M dataset, the models reach an accuracy of around 24\%. \textit{Even though the models trained with $\mathcal{L}^{pre}_{\text{MMCL}} + \mathcal{L}^{pre}_{\text{SSL}}$ attained higher accuracy than the models trained with $\mathcal{L}^{pre}_{\text{MMCL}}$ alone, in order to have better visualization of the difference in performance of CleanCLIP on the two pre-training objectives, we deliberately choose models with similar starting accuracies. } 
All the models, irrespective of the pre-training objective and the training dataset, reached more than 99\% ASR (see \Cref{tab:starting-accuracy-asr} in \Cref{sec:pre-training-details-and-starting-acc-asr}), implying that poisoning just 0.05\% of the dataset is enough to attain very high ASR.

\begin{table*}
    \centering
    \caption{This table shows the original Top-1 zero-shot Imagenet validation set accuracies and remaining accuracy after cleaning the models that were poisoned using BadNet by training from scratch. The cleaning is done using CleanCLIP, i.e., finetuning a poisoned model with $\mathcal{L}^{ft}_{\text{MMCL}} + \mathcal{L}^{ft}_{\text{SSL}}$. For this table, we choose the models having the highest accuracy and ASR $\le5\%$ (successful cleaning). The original ASR values for all models are more than 99\%. \textbf{Takeaway:} The models trained with $\mathcal{L}^{pre}_{\text{MMCL}}$ maintain their original accuracy after cleaning, while the ones trained with $\mathcal{L}^{pre}_{\text{MMCL}} + \mathcal{L}^{pre}_{\text{SSL}}$ experience a huge drop relative to the starting accuracy ($\sim$20\% for model trained on CC3M dataset and 45\% for model trained on CC6M dataset) after cleaning. }

    \begin{tikzpicture}[remember picture]
    \node (table) [inner sep=0pt] {
    \footnotesize
    \begin{tabular}{ccccccc}
    \toprule
    &   &   & \multicolumn{2}{c}{Trained with $\mathcal{L}^{pre}_{\text{MMCL}}$} & \multicolumn{2}{c}{Trained with $\mathcal{L}^{pre}_{\text{MMCL}} + \mathcal{L}^{pre}_{\text{SSL}}$} \\ \midrule
    Backdoor & Dataset & Clean Data Size &  Orig. Acc. &  Clean Acc. (ASR $\le$ 5\%) $\uparrow$ & Orig. Acc. & Clean Acc. (ASR $\le$ 5\%) $\uparrow$ \\ \cmidrule(lr){4-5} \cmidrule(lr){6-7}
    BadNet & CC3M     & 100K          &  \textbf{16.00\%}  \tikzmark{start1}    &  \tikzmark{end1} 16.49\% \tikzmark{emoji1} & \textbf{17.04\% } \tikzmark{start2}  & \tikzmark{end2} 14.16\% \tikzmark{emoji2} \vspace{1em} \\ 
    BadNet & CC6M   & 100K          &  \textbf{23.76\%}   \tikzmark{start3}  & \tikzmark{end3}  24.04\%    & \textbf{23.86\%}   \tikzmark{start4}  & \tikzmark{end4} 13.05\%   \\
    \bottomrule
    \end{tabular}
    };
    \end{tikzpicture}
    
    \begin{tikzpicture}[overlay, remember picture]
    \node at ([xshift=5mm, yshift=-2mm]pic cs:emoji1) {\includegraphics[height=2em]{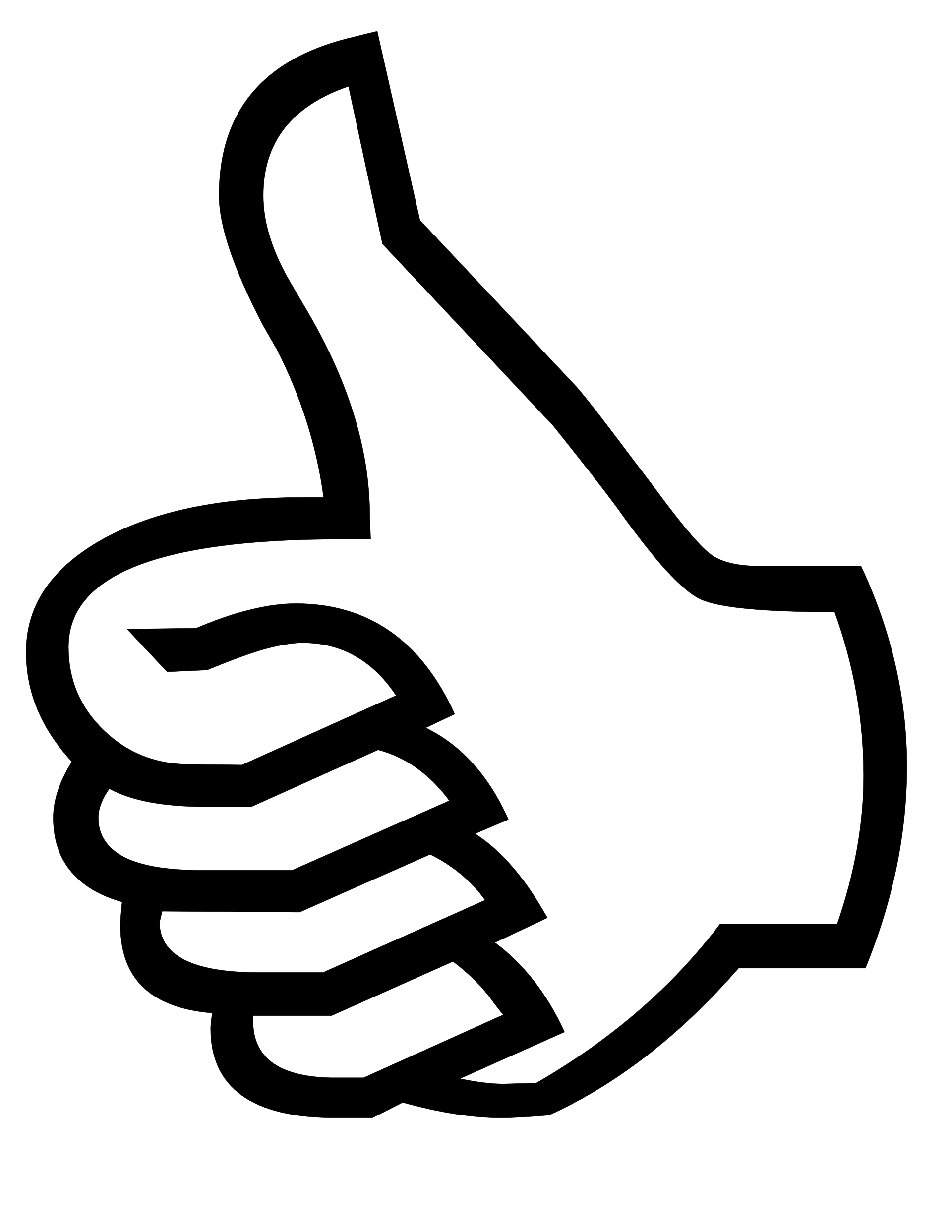}};
    \node at ([xshift=5mm, yshift=-2mm]pic cs:emoji2) {\includegraphics[height=2em]{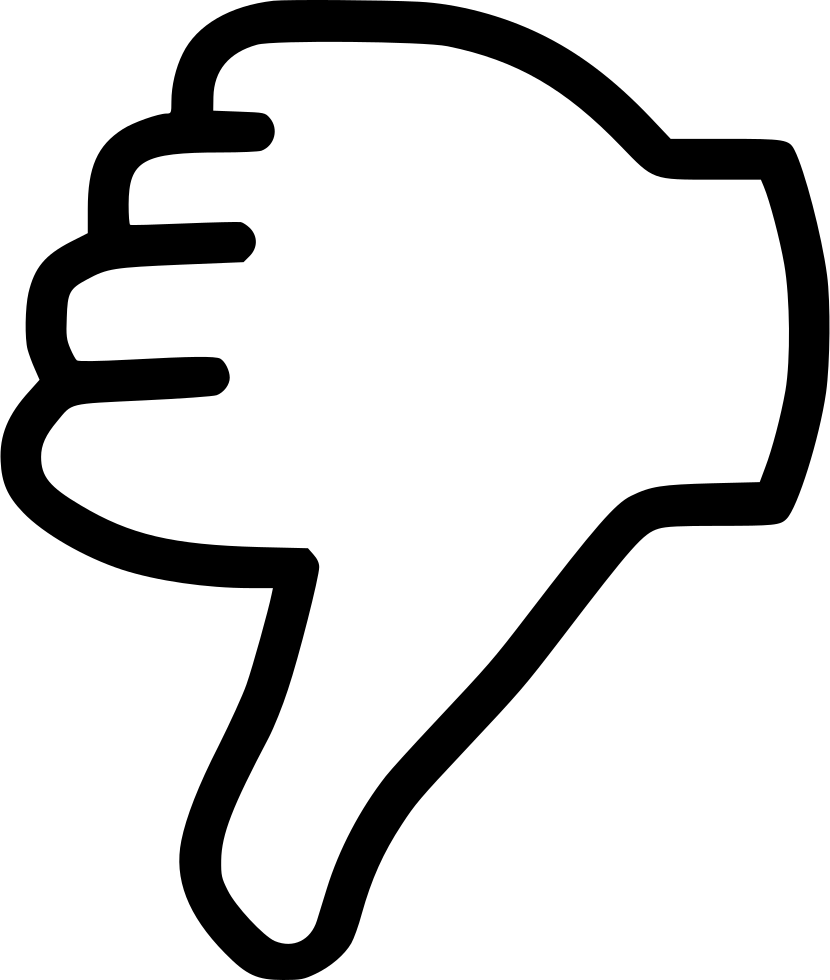}};
    \end{tikzpicture}
    \label{tab:start-final-accuracy}
    \begin{tikzpicture}[remember picture, overlay]
        \draw[->, thick, red] ([yshift=1mm]pic cs:start1) -- ([yshift=1mm]pic cs:end1);
    \end{tikzpicture}
    \begin{tikzpicture}[remember picture, overlay]
        \draw[->, thick, red] ([yshift=1mm]pic cs:start2) -- ([yshift=1mm]pic cs:end2);
    \end{tikzpicture}
    \begin{tikzpicture}[remember picture, overlay]
        \draw[->, thick, red] ([yshift=1mm]pic cs:start3) -- ([yshift=1mm]pic cs:end3);
    \end{tikzpicture}
    \begin{tikzpicture}[remember picture, overlay]
        \draw[->, thick, red] ([yshift=1mm]pic cs:start4) -- ([yshift=1mm]pic cs:end4);
    \end{tikzpicture}
\end{table*}

\paragraph{Removing Poison}
We clean the poisoned model by finetuning it on a 100K, guaranteed to be poison-free, image-text pairs for 20 epochs using a batch size of 128 and AdamW as the optimizer. We perform extensive hyperparameter search and use various learning rates (as many as 8 in some experiments and 14 in others, all with cosine scheduling and 50 warmup steps) for this process. Please refer to \Cref{sec:hyperparam-details-appendix} for the set of learning rates explored for the cleaning procedure. 
Ideally, after cleaning we would want to obtain a model that maintains the accuracy of the original poisoned model, while getting rid of its poison, i.e., very low ASR. 
We use three different loss functions for the cleaning process: 
\begin{enumerate}[leftmargin=7pt,topsep=0pt,itemsep=2pt]
    \item $\mathcal{L}^{ft}_{\text{MMCL}}$: CleanCLIP showed that finetuning with $\mathcal{L}^{ft}_{\text{MMCL}}$ did not change the original model's accuracy and ASR, and hence is an \textit{ineffective cleaning loss}. We reproduce these results for the models we trained. 
    
    \item $\mathcal{L}^{ft}_{\text{SSL}}$: CleanCLIP also showed that finetuning with $\mathcal{L}^{ft}_{\text{SSL}}$ decreased the original model's ASR at the expense of its accuracy, and hence is also an \textit{ineffective cleaning loss}. We also reproduce these results. 
    
    \item $\mathcal{L}^{ft}_{\text{MMCL}} + \mathcal{L}^{ft}_{\text{SSL}}$: CleanCLIP showed that finetuning with a combination of these losses decreased the original model's ASR while not hurting its accuracy, and hence is an \textit{effective cleaning loss}. Our experiments show that while this observation is true for the models trained with $\mathcal{L}^{pre}_{\text{MMCL}}$, however \textit{it does not generalize} to the models trained with stronger pre-training objective $\mathcal{L}^{pre}_{\text{MMCL}} + \mathcal{L}^{pre}_{\text{SSL}}$. \textbf{This is our key finding. } 
\end{enumerate}

\begin{figure*}
    \centering
    \begin{tikzpicture}
        \node[anchor=south west,inner sep=0] (image) at (0,0) {
            \includegraphics[width=0.9\textwidth]{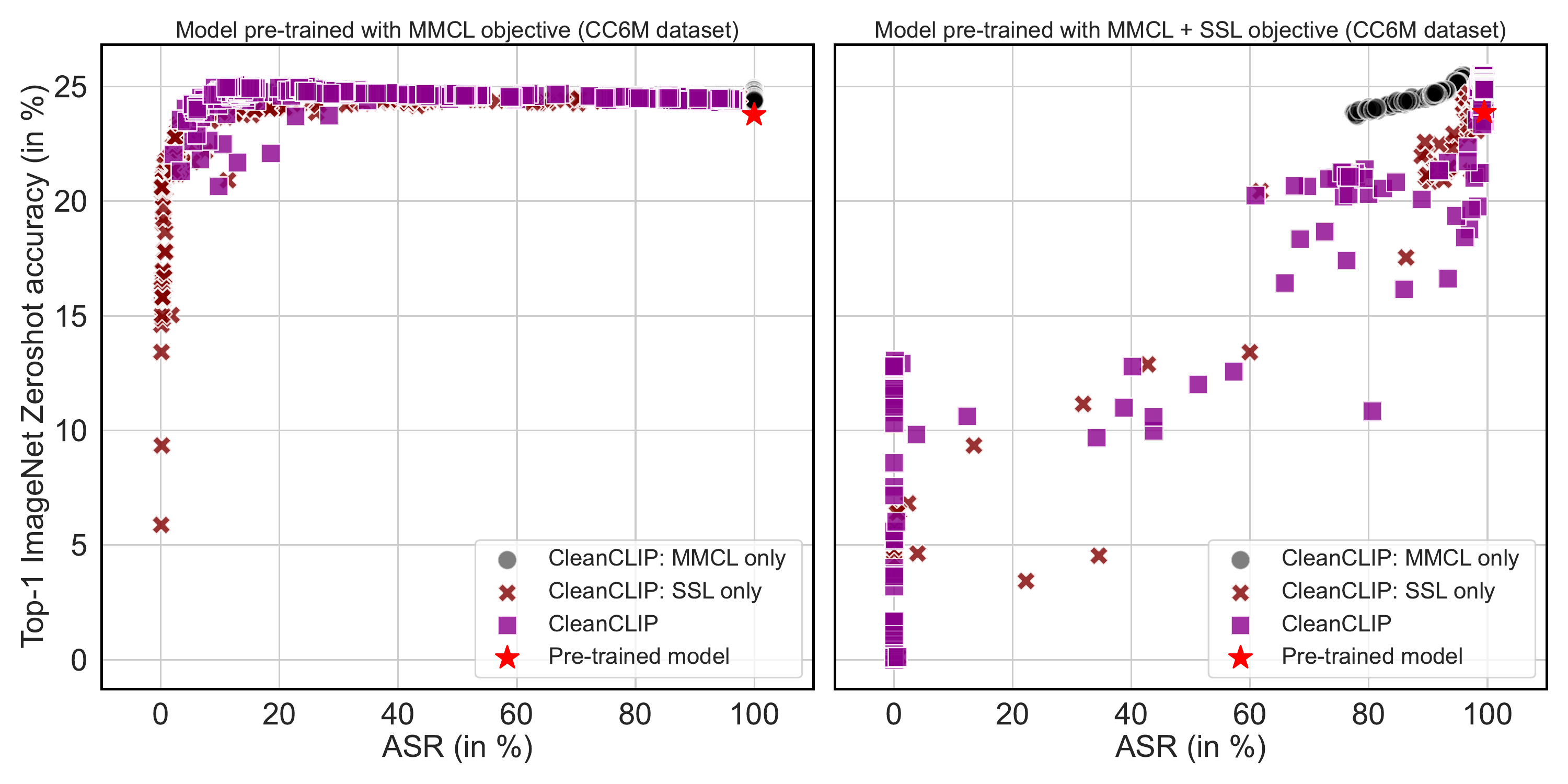}
        };
        \begin{scope}[x={(image.south east)},y={(image.north west)}]
            \draw[red, thick] (0.11,0.865) ellipse (0.06*0.5 and 0.06*1);
            \draw[red, thick] (0.57,0.865) ellipse (0.06*0.5 and 0.06*1);
            \draw[-{Stealth[length=2mm,width=2mm]}, red, thick] (0.045,0.98) -- (0.086,0.905);
            \draw[-{Stealth[length=2mm,width=2mm]}, red, thick] (0.51,0.98) -- (0.55,0.905);

            \draw[black, thick] (0.485,0.855) ellipse (0.03*0.5 and 0.03*1);
            \draw[black, thick] (0.945,0.855) ellipse (0.03*0.5 and 0.03*1);
            \draw[-{Stealth[length=2mm,width=2mm]}, black, thick] (0.525,0.945) -- (0.49,0.88);
            \draw[-{Stealth[length=2mm,width=2mm]}, black, thick] (1,0.95) -- (0.95,0.881);

            \draw[red, thick] (0.57, 0.865) -- (0.57, 0.55); 
            \draw[red, thick] (0.56, 0.865) -- (0.58, 0.865); 
            \draw[red, thick] (0.56, 0.55) -- (0.58, 0.55); 
            
            \node[right, text=red] at (0.57, 0.70) {45\% drop}; 
        \end{scope}
      \end{tikzpicture}

    \caption{Top-1 zero-shot Imagenet validation set accuracy vs. the ASR, measured at the end of each cleaning epoch for the models trained on the CC6M dataset. The cleaning is done by finetuning the model with the three losses mentioned above. The red star in the top right corner (encircled in the black circle) corresponds to the model's starting accuracy and ASR (before cleaning). For a successful cleaning, there should be models that maintain the model's starting accuracy while having a low ASR (indicated by the red circle's region in the top left). There are several models in the red circle in the left plot (successful clean), while there are no models in the red circle in the right plot (unsuccessful clean). 
    \textbf{Takeaway:} CleanCLIP successfully cleans the model trained with $\mathcal{L}^{pre}_{\text{MMCL}}$ (left), while it is ineffective for the models trained with $\mathcal{L}^{pre}_{\text{MMCL}} + \mathcal{L}^{pre}_{\text{SSL}}$ (right).}
    \label{fig:CC6M-cleaning-plots}
\end{figure*}

\paragraph{Findings from the Cleaning Procedure}
\Cref{fig:CC6M-cleaning-plots} shows the scatter plot of the Top-1 zero-shot Imagenet validation set accuracy and the ASR at the end of each cleaning epoch for the models trained on the CC6M dataset. We defer the plots for the CC3M dataset in \Cref{sec:cc3m-main-results-appendix} for space consideration. 
For both the datasets, we observe that:
\begin{enumerate}[leftmargin=7pt,topsep=0pt,itemsep=2pt]
    \item $\mathcal{L}^{ft}_{\text{MMCL}}$ and $\mathcal{L}^{ft}_{\text{SSL}}$ individually are ineffective cleaning losses as they cause a significant drop in accuracy for lowering the ASR for both the pre-training objectives. 

    \item $\mathcal{L}^{ft}_{\text{MMCL}} + \mathcal{L}^{ft}_{\text{SSL}}$ serves as an effective cleaning loss for the model trained with $\mathcal{L}^{pre}_{\text{MMCL}}$ (left plot). The cleaned models maintain the accuracy of the original model, but they have low ASR, which we consider \emph{successful cleaning}. 
    However, it does not lead to an effective cleaning of the model trained with $\mathcal{L}^{pre}_{\text{MMCL}} + \mathcal{L}^{pre}_{\text{SSL}}$ (right plot). Even the model that has the highest accuracy with a low ASR ($\le$ 5\%) is 45\% less accurate than the original model, as shown in \Cref{fig:CC6M-cleaning-plots}. 
\end{enumerate}

For both datasets, our findings indicate that CleanCLIP is not effective in removing poison from the models trained with a stronger pre-training objective $\mathcal{L}^{pre}_{\text{MMCL}} + \mathcal{L}^{pre}_{\text{SSL}}$, without a significant drop in accuracy. \Cref{tab:start-final-accuracy} gives the highest accuracy of the models which were successfully cleaned by CleanCLIP (ASR $\le$ 5\%). 

\paragraph{Poison Induction by Finetuning a pre-trained model}
We also induce poison by finetuning a pre-trained CLIP model \citep{Radford2019CLIP}. Concretely, we poison two models by finetuning them with $\mathcal{L}^{pre}_{\text{MMCL}}$ and $\mathcal{L}^{pre}_{\text{MMCL}} + \mathcal{L}^{pre}_{\text{SSL}}$ respectively, using the CC6M dataset that had 3000 poisoned datapoints. We use a learning rate of $5e-7$ with AdamW optimizer with 10,000 warmup steps. After poisoning, these models achieve Top-1 zero-shot Imagenet set accuracy of $\sim60\%$, much higher than the models trained from scratch (\Cref{fig:CC6M-cleaning-plots}). The ASR for the model poisoned with $\mathcal{L}^{pre}_{\text{MMCL}}$ is 99\% and for the model poisoned with $\mathcal{L}^{pre}_{\text{MMCL}} + \mathcal{L}^{pre}_{\text{SSL}}$ is 90\%. 

\begin{figure*}
    \centering
    \begin{tikzpicture}
        \node[anchor=south west,inner sep=0] (image) at (0,0) {
            \includegraphics[width=0.9\textwidth]{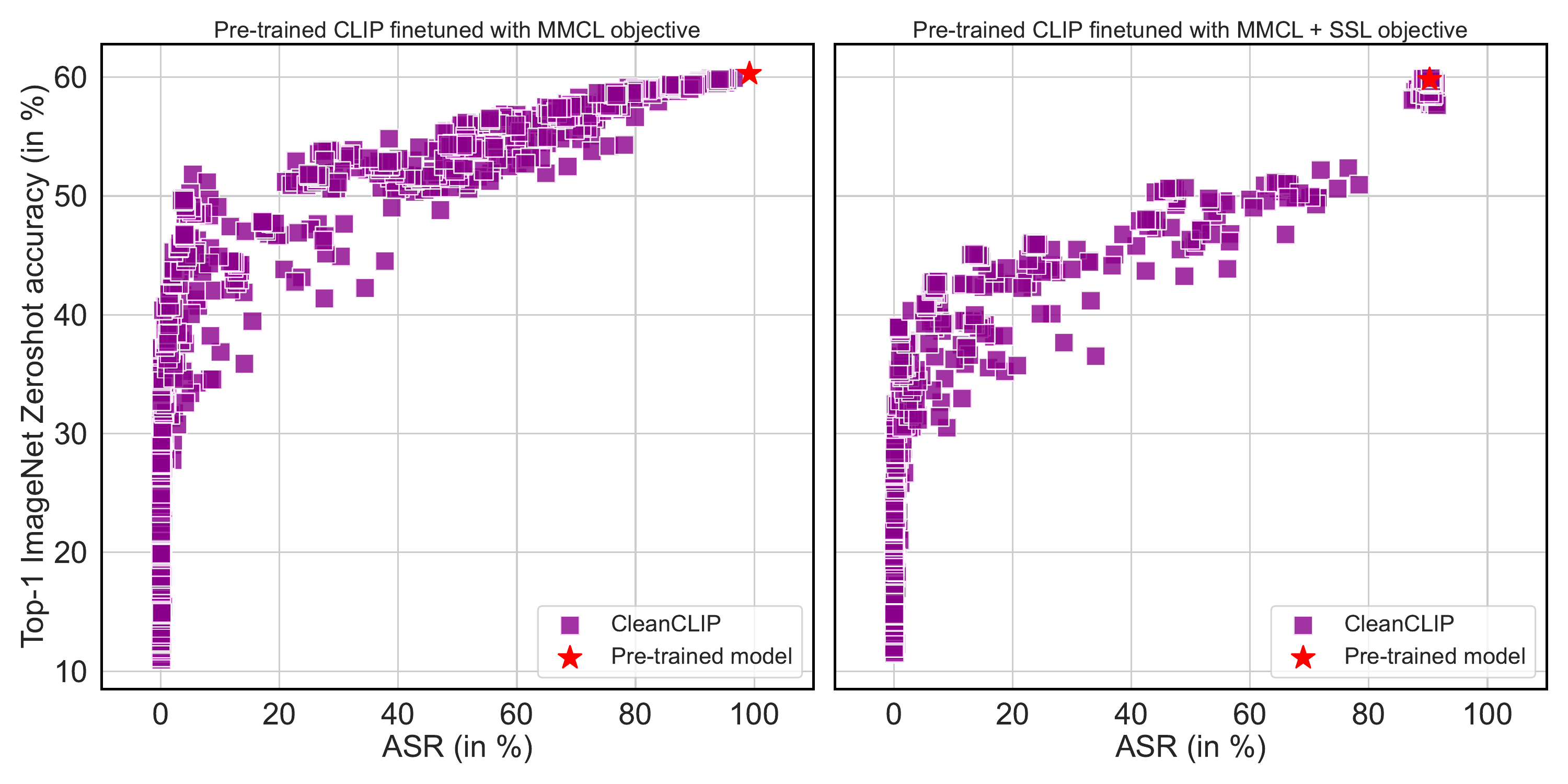}
        };
        \begin{scope}[x={(image.south east)},y={(image.north west)}]
            \draw[red, thick] (0.11,0.865) ellipse (0.06*0.5 and 0.06*1);
            \draw[red, thick] (0.57,0.865) ellipse (0.06*0.5 and 0.06*1);
            \draw[-{Stealth[length=2mm,width=2mm]}, red, thick] (0.045,0.98) -- (0.086,0.905);
            \draw[-{Stealth[length=2mm,width=2mm]}, red, thick] (0.52,0.97) -- (0.55,0.905);

            \draw[black, thick] (0.48,0.905) ellipse (0.03*0.5 and 0.03*1);
            \draw[black, thick] (0.91,0.895) ellipse (0.03*0.5 and 0.03*1);
            \draw[-{Stealth[length=2mm,width=2mm]}, black, thick] (0.53,1.0) -- (0.49,0.93);
            \draw[-{Stealth[length=2mm,width=2mm]}, black, thick] (0.975,0.98) -- (0.925,0.911);

            \draw[red, thick] (0.11, 0.9) -- (0.11, 0.75); 
            \draw[red, thick] (0.10, 0.9) -- (0.12, 0.9); 
            \draw[red, thick] (0.10, 0.75) -- (0.12, 0.75); 
            \node[right, text=red] at (0.105, 0.85) {17\% drop}; 
            
            \draw[red, thick] (0.57, 0.9) -- (0.57, 0.58); 
            \draw[red, thick] (0.56, 0.9) -- (0.58, 0.9); 
            \draw[red, thick] (0.56, 0.58) -- (0.58, 0.58); 
            \node[right, text=red] at (0.57, 0.75) {33\% drop}; 
        \end{scope}
      \end{tikzpicture}
    \caption{Top-1 zero-shot Imagenet validation set accuracy vs. the ASR, measured at the end of each cleaning epoch for the models poisoned by finetuning a CLIP pre-trained checkpoint on the CC6M dataset. The cleaning is done by finetuning the poisoned model with $\mathcal{L}^{ft}_{\text{MMCL}} + \mathcal{L}^{ft}_{\text{SSL}}$. The red star in the top right corner (encircled in the black circle) corresponds to the original model's accuracy and ASR (before cleaning). For a successful cleaning, there should be models that maintain the original model's accuracy while having a low ASR (indicated by the red circle in the top left). \textbf{Takeaway:} CleanCLIP is unable to successfully clean both the models; however, it performs much worse for the model poisoned with $\mathcal{L}^{pre}_{\text{MMCL}} + \mathcal{L}^{pre}_{\text{SSL}}$ (right). }
    \label{fig:finetuning-poison-CC6M-cleaning-plots}
\end{figure*}

\textbf{Findings from the Cleaning Procedure:} We clean the poisoned models using CleanCLIP, i.e., further finetuning on a clean dataset with $\mathcal{L}^{ft}_{\text{MMCL}} + \mathcal{L}^{ft}_{\text{SSL}}$. \Cref{fig:finetuning-poison-CC6M-cleaning-plots} shows the scatter plot of the Top-1 zero-shot Imagenet validation set accuracy and the ASR at the end of each finetuning epoch for these two models. In this case, both the models experience a drop in accuracy to obtain a low ASR ($\le$ 5\%); however, the drop is much higher for the model when the poison was induced using $\mathcal{L}^{pre}_{\text{MMCL}} + \mathcal{L}^{pre}_{\text{SSL}}$ (33\%, compared to a 17\% drop for the model when the poison was induced using $\mathcal{L}^{pre}_{\text{MMCL}}$). This experiment corroborates our previous finding that CleanCLIP is less effective when the poison is induced using a combination of MMCL and SSL, irrespective of the fact whether the poison is induced via finetuning or by training from scratch. 

\paragraph{Poisoning a Different Backbone Architecture}

\begin{figure*}
    \centering
    \begin{tikzpicture}
        \node[anchor=south west,inner sep=0] (image) at (0,0) {
            \includegraphics[width=0.9\textwidth]{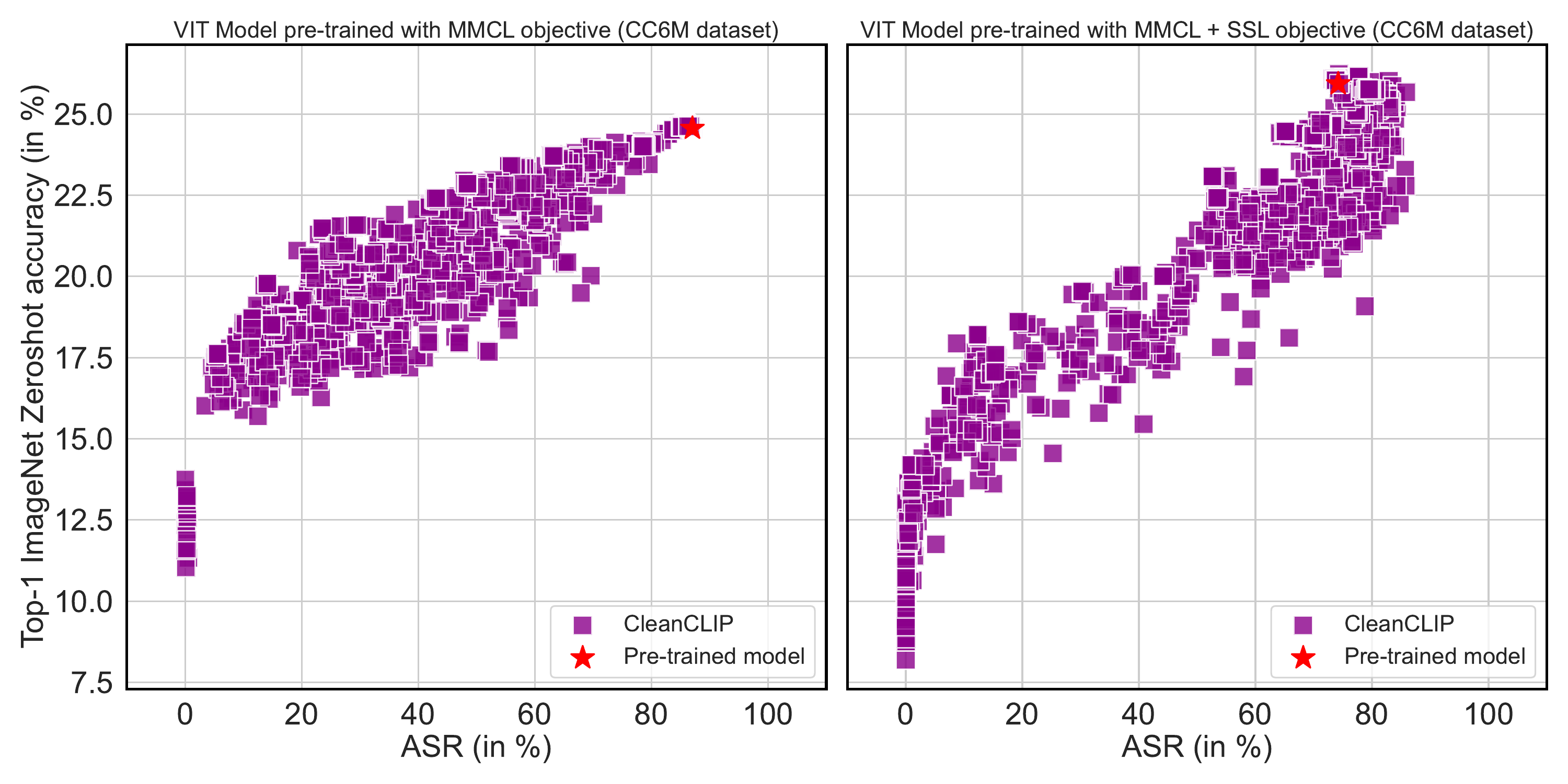}
        };
        \begin{scope}[x={(image.south east)},y={(image.north west)}]
            \draw[red, thick] (0.11,0.845) ellipse (0.05*0.5 and 0.05*1);
            \draw[red, thick] (0.57,0.885) ellipse (0.05*0.5 and 0.05*1);
            \draw[-{Stealth[length=2mm,width=2mm]}, red, thick] (0.045,0.96) -- (0.086,0.88);
            \draw[-{Stealth[length=2mm,width=2mm]}, red, thick] (0.52,0.97) -- (0.55,0.915);

            \draw[black, thick] (0.44,0.84) ellipse (0.03*0.5 and 0.03*1);
            \draw[black, thick] (0.85,0.895) ellipse (0.03*0.5 and 0.03*1);
            \draw[-{Stealth[length=2mm,width=2mm]}, black, thick] (0.49,0.94) -- (0.45,0.87);
            \draw[-{Stealth[length=2mm,width=2mm]}, black, thick] (0.915,0.98) -- (0.865,0.911);

            \draw[red, thick] (0.11, 0.845) -- (0.11, 0.53); 
            \draw[red, thick] (0.10, 0.845) -- (0.12, 0.845); 
            \draw[red, thick] (0.10, 0.53) -- (0.12, 0.53); 
            \node[right, text=red] at (0.105, 0.74) {30\% drop}; 
            
            \draw[red, thick] (0.57, 0.89) -- (0.57, 0.46); 
            \draw[red, thick] (0.56, 0.89) -- (0.58, 0.89); 
            \draw[red, thick] (0.56, 0.46) -- (0.58, 0.46); 
            \node[right, text=red] at (0.57, 0.74) {41\% drop}; 
        \end{scope}
      \end{tikzpicture}
    \caption{Top-1 zero-shot Imagenet validation set accuracy vs. the ASR, measured at the end of each cleaning epoch for the models trained on the CC6M dataset. The cleaning using CleanCLIP. The red star in the top right corner (encircled in the black circle) corresponds to the model's starting accuracy and ASR (before cleaning). \textbf{Takeaway:} When a ViT backbone is poisoned, there are no cleaned models that maintain the original accuracies for both the pre-training losses, however the drop is much larger for the model trained with $\mathcal{L}^{pre}_{\text{MMCL}} + \mathcal{L}^{pre}_{\text{SSL}}$ (right). }
    \label{fig:ViT-from-scratch-poison-CC6M-cleaning-plots}
\end{figure*}

For this experiment, we poison a model with a different backbone architecture, specifically ViTs. We poisoned these models by training them on the CC6M dataset with 3000 poisoned datapoints, and cleaned them using CleanCLIP. \Cref{fig:ViT-from-scratch-poison-CC6M-cleaning-plots} shows the cleaning plots for these models. We observe that similar to the previous experiments, the model trained with MMCL + SSL experiences a much larger drop in accuracy than the model just trained with MMCL, although in this case the model just trained with MMCL experiences a lot drop in accuracy as well.

\paragraph{Poisoning Induction using a Different Poison}
Due to space considerations, we present the results for the effectiveness of CleanCLIP when models are poisoned using label consistent backdoors in \Cref{sec:label-consistent-poison-removal}. We observe that similar to the case of poisoning with BadNet, CleanCLIP is much less effective when the poison is induced using $\mathcal{L}^{pre}_{\text{MMCL}} + \mathcal{L}^{pre}_{\text{SSL}}$ (16\% loss in accuracy), compared to the case when it is induced using $\mathcal{L}^{pre}_{\text{MMCL}}$ (10\% gain in accuracy).

\begin{figure}[h]
    \centering
    \begin{subfigure}{\textwidth}
    \centering
    \begin{tikzpicture}
        \node[anchor=south west,inner sep=0] (image) at (0,0) {
            \includegraphics[width=0.9\textwidth]{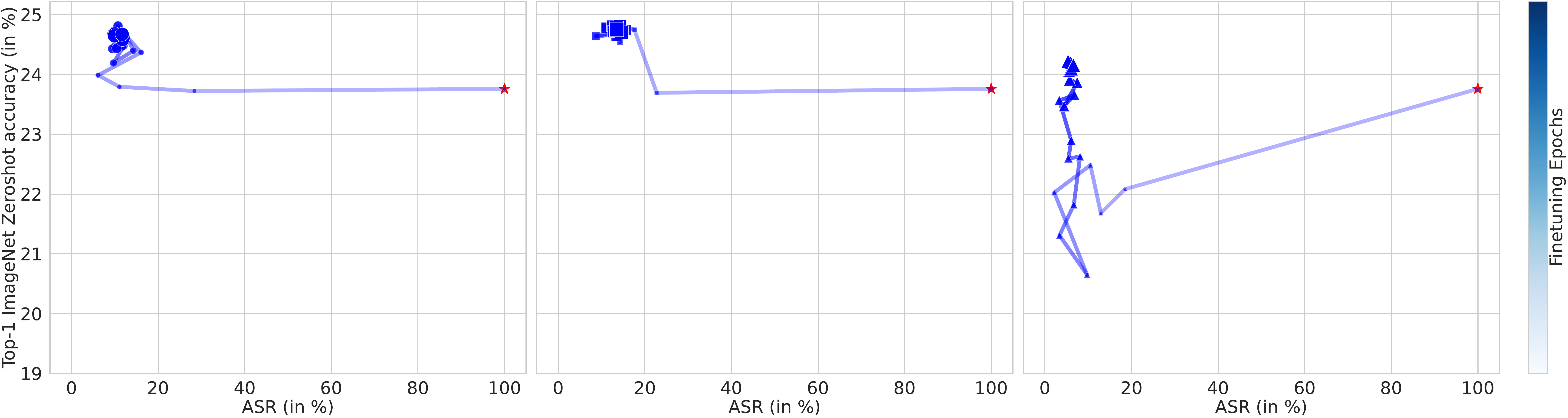}
        };
        \begin{scope}[x={(image.south east)},y={(image.north west)}]
            \draw[black, thick] (0.32,0.79) ellipse (0.03*0.5 and 0.03*1.7);
            \draw[black, thick] (0.635,0.79) ellipse (0.03*0.5 and 0.03*1.7);
            \draw[black, thick] (0.94,0.79) ellipse (0.03*0.5 and 0.03*1.7);
            \draw[-{Stealth[length=2mm,width=2mm]}, black, thick] (0.35,0.9) -- (0.33,0.82);
            \draw[-{Stealth[length=2mm,width=2mm]}, black, thick] (0.665,0.9) -- (0.645,0.82);
            \draw[-{Stealth[length=2mm,width=2mm]}, black, thick] (0.97,0.9) -- (0.95,0.82);
        \end{scope}
      \end{tikzpicture}
        \caption{Finetuning trajectory of three runs with different learning rates for the model pre-trained with $\mathcal{L}^{pre}_{\text{MMCL}}$. The trajectories are smooth and end up in high accuracy and low ASR regime at the end of finetuning process. }
        \label{fig:mmcl-finetuning-trajectory}    
    \end{subfigure}

    \begin{subfigure}{\textwidth}
    \centering
        \begin{tikzpicture}
        \node[anchor=south west,inner sep=0] (image) at (0,0) {
            \includegraphics[width=0.9\textwidth]{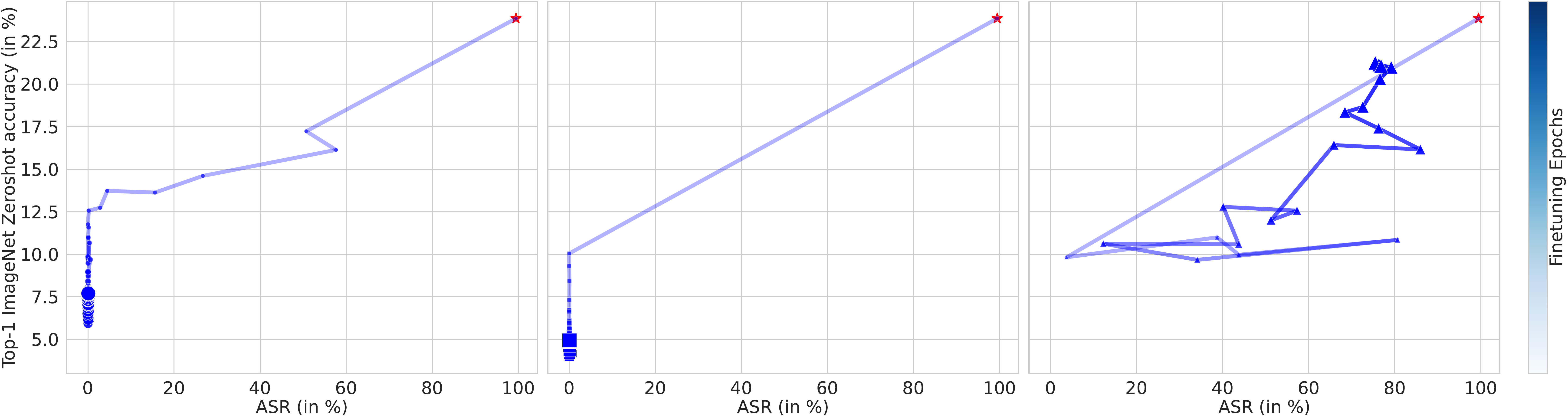}
        };
        \begin{scope}[x={(image.south east)},y={(image.north west)}]
            \draw[black, thick] (0.33,0.94) ellipse (0.03*0.5 and 0.03*1.7);
            \draw[black, thick] (0.635,0.94) ellipse (0.03*0.5 and 0.03*1.7);
            \draw[black, thick] (0.94,0.94) ellipse (0.03*0.5 and 0.03*1.7);
            \draw[-{Stealth[length=2mm,width=2mm]}, black, thick] (0.37,1.03) -- (0.34,0.97);
            \draw[-{Stealth[length=2mm,width=2mm]}, black, thick] (0.675,1.03) -- (0.645,0.97);
            \draw[-{Stealth[length=2mm,width=2mm]}, black, thick] (0.98,1.03) -- (0.95,0.97);
        \end{scope}
      \end{tikzpicture}
        \caption{Finetuning trajectory of three runs with different learning rates for the model pre-trained with $\mathcal{L}^{pre}_{\text{MMCL}} + \mathcal{L}^{pre}_{\text{SSL}}$. The trajectories are not smooth and increased finetuning can lead to both decreased accuracy and higher ASR.}
        \label{fig:mmcl-ssl-finetuning-trajectory}
    \end{subfigure}
    \vspace{1pt}
    \caption{Finetuning trajectories of models with different pre-training objectives. Successive finetuning epochs are shown with increasing size of the markers and intensity of the connecting line. The red star in the top right corner (encircled in the black circle) corresponds to the original model's accuracy and ASR. \textbf{Takeaway:} Models trained with $\mathcal{L}^{pre}_{\text{MMCL}}$ converge to a region of high accuracy and low ASR as we continue to finetune. On the other hand, models trained with $\mathcal{L}^{pre}_{\text{MMCL}} + \mathcal{L}^{pre}_{\text{SSL}}$ fail to converge to a region of high accuracy and low ASR, and continued finetuning can lead to both decreased accuracy and higher ASR. This makes determining the stopping criterion for the cleaning process for the latter models challenging. }
    \label{fig:finetuning-trajectory-combined}
\end{figure}

\paragraph{Dependence of the Stopping Criterion on the Pre-training Objective}
In the previous section, the ability to find a model with high accuracy and low ASR is considered a success for the CleanCLIP approach. However, in practice, one would not be aware of the ASR of the model being cleaned and, therefore, would not know when to stop the cleaning process. To highlight this practicality concern, we show the multiple cleaning trajectories for models trained using different pre-training objectives in \Cref{fig:finetuning-trajectory-combined}. 

\Cref{fig:mmcl-finetuning-trajectory} shows the trajectories of three cleaning runs with different learning rates for a model trained using $\mathcal{L}^{pre}_{\text{MMCL}}$ on the CC6M dataset. 
We observe that in all the three runs, the trajectory converges to a region of high accuracy and low ASR (top left corner), and the trajectories are smooth. This indicates that a practitioner can clean this model by finetuning the poisoned model for as long as their resources allow, and choose the model at the end of the finetuning process. They will likely obtain a model with high accuracy and low ASR, i.e., a successfully cleaned model. 

\Cref{fig:mmcl-ssl-finetuning-trajectory} shows the trajectories of three cleaning runs with different learning rates for a model trained using $\mathcal{L}^{pre}_{\text{MMCL}} + \mathcal{L}^{pre}_{\text{SSL}}$ on the CC6M dataset. 
We observe that in all three runs, the successfully cleaned model (high accuracy with a low ASR ($\le$ 5\%)) is an intermediate model in the trajectory, \textit{and not} the model at the end of the process. With continued finetuning, the model can both lose accuracy and gain ASR, both of which are undesirable. Therefore, it is difficult for a practitioner to discern when to stop the finetuning process to obtain a clean model. 

Therefore, the practicality of using CleanCLIP also depends on the pre-training objective of the model. 
\Cref{sec:cleaning-trajectories-appendix} shows the cleaning trajectories for all explored hyperparameters.

\paragraph{Dependence of CleanCLIP on the Ideal Condition of the Dataset}
CleanCLIP assumes that the cleaning data is entirely free of poisoned datapoints. In practice, this assumption can be violated even when considerable care is taken to ensure it. To simulate this real-world situation, we clean models using data with a few poisoned datapoints, specifically 5 and 10 poisoned datapoints in the 100K cleaning datapoints. 
Note that these datasets are still, respectively, 10$\times$ and 5$\times$ cleaner than the original training dataset, illustrating a situation where the cleaning data is much cleaner than the training dataset but still not perfect.

\begin{figure}
    \centering
    \begin{tikzpicture}
        \node[anchor=south west,inner sep=0] (image) at (0,0) {
            \includegraphics[width=0.9\textwidth]{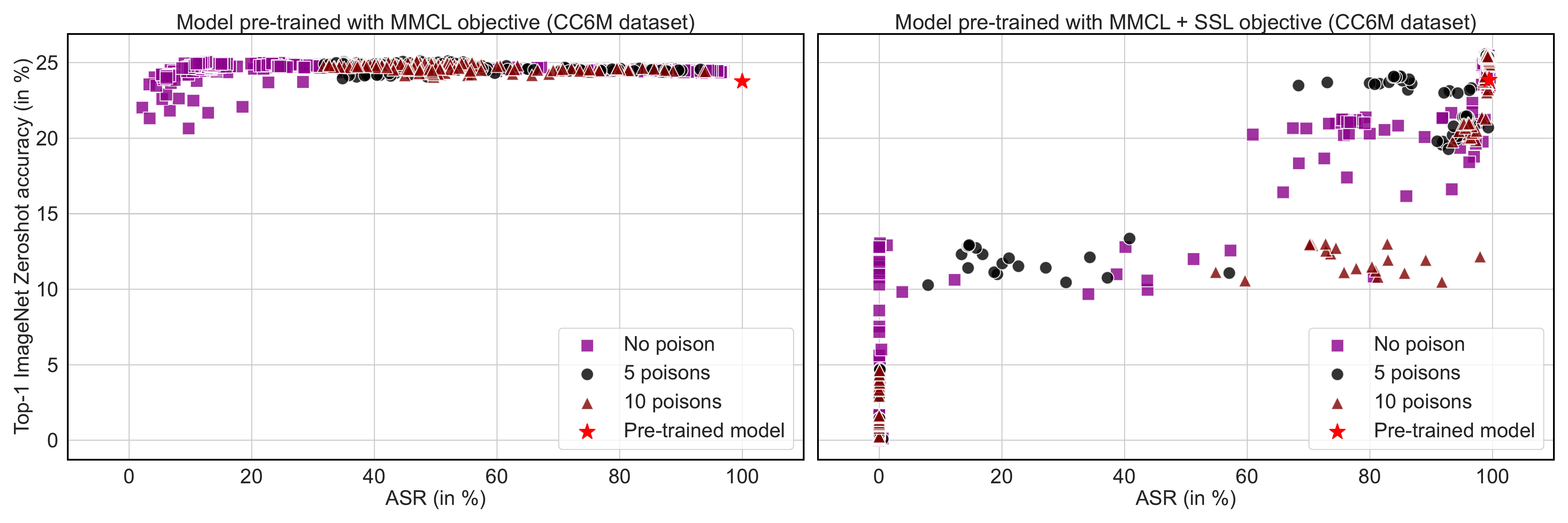}
        };
        \begin{scope}[x={(image.south east)},y={(image.north west)}]
            \draw[red, thick] (0.09,0.86) ellipse (0.06*0.33 and 0.06*1);
            \draw[red, thick] (0.56,0.86) ellipse (0.06*0.33 and 0.06*1);
            
            \draw[black, thick] (0.475,0.845) ellipse (0.04*0.33 and 0.04*1);
            \draw[black, thick] (0.95,0.85) ellipse (0.04*0.33 and 0.04*1);
            
            \draw[-{Stealth[length=2mm,width=2mm]}, red, thick] (0.045,0.98) -- (0.076,0.90);
            \draw[-{Stealth[length=2mm,width=2mm]}, red, thick] (0.515,0.98) -- (0.546,0.90);
            
            \draw[-{Stealth[length=2mm,width=2mm]}, black, thick] (0.515,0.925) -- (0.485,0.865);
            \draw[-{Stealth[length=2mm,width=2mm]}, black, thick] (0.99,0.93) -- (0.96,0.87);
        \end{scope}
      \end{tikzpicture}

    \caption{Top-1 zero-shot Imagenet validation set accuracy v/s the ASR during the cleaning process for the two models. The finetuning is done with $\mathcal{L}^{ft}_{\text{MMCL}} + \mathcal{L}^{ft}_{\text{SSL}}$. We measure accuracy and ASR at the end of each epoch. The red star in the top right corner (encircled in the black circle) corresponds to the original model's accuracy and ASR. For a successful cleaning, there should be models that maintain the original model's accuracy while having a low ASR (indicated by the red circle). \textbf{Takeaway:} Even having 5 poisons in the cleaning dataset (i.e. 0.005\% of the dataset, which is 10$\times$ cleaner than the pre-training data) hurts the cleaning process for both pre-training objectives, and $\mathcal{L}^{pre}_{\text{MMCL}} + \mathcal{L}^{pre}_{\text{SSL}}$ trained models are hurt worse. }
    \label{fig:CC6M-cleaning-plots-slight-poison}
\end{figure}

\Cref{fig:CC6M-cleaning-plots-slight-poison} shows the scatter plot of the Top-1 zero-shot Imagenet validation set accuracy and the ASR at the end of each cleaning epoch when two models trained from scratch on the CC6M dataset, one using $\mathcal{L}^{pre}_{\text{MMCL}}$ and the other using $\mathcal{L}^{pre}_{\text{MMCL}} + \mathcal{L}^{pre}_{\text{SSL}}$ is cleaned by finetuning on this slightly poisoned dataset with $\mathcal{L}^{ft}_{\text{MMCL}} + \mathcal{L}^{ft}_{\text{SSL}}$. 
We observe that having just 5 poisoned datapoints in the cleaning dataset severely weakens CleanCLIP for both the pre-training objectives. 

For models pre-trained with $\mathcal{L}^{pre}_{\text{MMCL}}$, we found cleaned models that maintain the original model's accuracy and achieve around 30-50\% ASR. On the other hand, for the models pre-trained with the stronger objective $\mathcal{L}^{pre}_{\text{MMCL}} + \mathcal{L}^{pre}_{\text{SSL}}$, having just 5 poisoned examples renders the cleaning procedure completely ineffective. The models pre-trained on the CC6M dataset lose about 80\% of the original model's accuracy to obtain a low ASR ($\le$ 5\%), and no model has an ASR lower than 90\% for the CC3M pre-trained model (\Cref{fig:CC3M-cleaning-plots-slight-poison}). 

\paragraph{Takeaways} Our experiments highlight the fact that a stronger pre-training objective, like the combination of MMCL and SSL, also affects the strength of poison induction, making the cleaning process difficult. 
Also, for a practitioner, when pre-training with a stronger objective, the decision of when to stop finetuning becomes non-trivial, as we show that the model at the end of the cleaning procedure is usually not the one with the lowest ASR and the best accuracy. 
The situation is further exacerbated when we even slightly relax the assumption of 100\% poison-free cleaning data, which can be too stringent in practice. 


\section{Analysis of the Stronger Pre-training Objective}
We now perform several analysis experiments to understand the reason behind the difference in the poison removal ability of CleanCLIP across the two pre-training objectives. We also experiment with other methods to attempt to remove the poison when induced using $\mathcal{L}^{pre}_{\text{MMCL}} + \mathcal{L}^{pre}_{\text{SSL}}$.

\paragraph{Cleaning using an Objective distinct from Pre-training}
CleanCLIP successfully cleans the models trained with $\mathcal{L}^{pre}_{\text{MMCL}}$ by finetuning with $\mathcal{L}^{ft}_{\text{MMCL}} + \mathcal{L}^{ft}_{\text{SSL}}$. However, it was unsuccessful for the model trained with $\mathcal{L}^{pre}_{\text{MMCL}} + \mathcal{L}^{pre}_{\text{SSL}}$. A plausible reason for this behavior could be that we are using the same pre-training and cleaning objective in the latter case, and CleanCLIP might be able to successfully clean the latter models if we were to clean it with a loss objective that is distinct from its pre-training objective. 
To test this hypothesis, we clean the models trained with $\mathcal{L}^{pre}_{\text{MMCL}} + \mathcal{L}^{pre}_{\text{SSL}}$ by finetuning with $\mathcal{L}^{ft}_{\text{MMCL}} + \mathcal{L}^{ft}_{\text{SSL}} + \mathcal{L}^{ft}_{\text{DeepClust}}$, where $\mathcal{L}^{ft}_{\text{DeepClust}}$ is an additional deep clustering objective \citep{deep-clustering-paper} on the vision encoder.

In deep clustering, we first obtain a pseudo-label for each image. We obtain the pseudo-label for an image in two ways: \textbf{a)} by classifying each image into one of the 1,000 Imagenet classes using powerful models such as SigLIP ViT-L/14 (zero-shot Imagenet accuracy of 83.08\%) \citep{siglip}, and \textbf{b)} performing a 1,000-way clustering on feature space of our trained vision encoder, using FAISS \citep{faiss}. Note that the use of SigLIP ViT-L/14 for obtaining pseudo-labels is a \textit{cheating experiment} for a deep clustering task (we don't have access to ground truth labels and therefore use SigLIP ViT-L/14 for this task); however, this experiment is solely performed to probe the upper bound of the poison removal that can be obtained when using deep clustering approaches.

\begin{figure}[h]
    \centering
    \begin{tikzpicture}
        \node[anchor=south west,inner sep=0] (image) at (0,0) {
            \includegraphics[width=0.5\columnwidth]{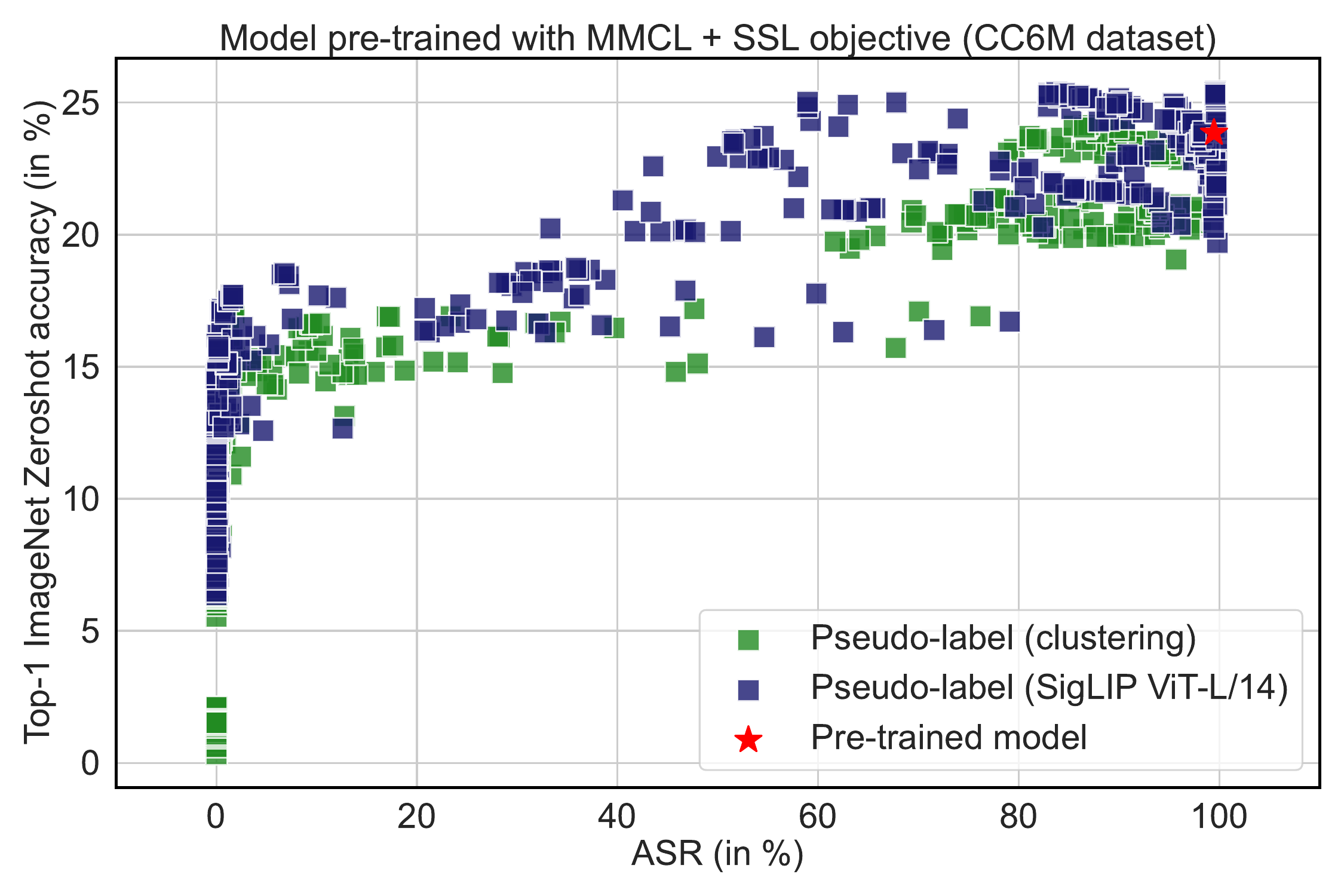}
        };
        \begin{scope}[x={(image.south east)},y={(image.north west)}]
            \draw[red, thick] (0.16,0.85) ellipse (0.06*0.66 and 0.06*1);
            \draw[black, thick] (0.9,0.85) ellipse (0.04*0.66 and 0.04*1);
            
            \draw[-{Stealth[length=2mm,width=2mm]}, red, thick] (0.08,0.96) -- (0.132,0.893);
            \draw[-{Stealth[length=2mm,width=2mm]}, black, thick] (0.97,0.95) -- (0.92,0.87);
        \end{scope}
      \end{tikzpicture}
    \caption{Top-1 zero-shot Imagenet validation set accuracy vs. the ASR during the cleaning process for the $\mathcal{L}^{pre}_{\text{MMCL}} + \mathcal{L}^{pre}_{\text{SSL}}$ pre-trained model on the CC6M dataset. The finetuning is done with $\mathcal{L}^{ft}_{\text{MMCL}} + \mathcal{L}^{ft}_{\text{SSL}} + \mathcal{L}^{ft}_{\text{DeepClust}}$. \textbf{Takeaway:} Adding $\mathcal{L}^{ft}_{\text{DeepClust}}$ is unable to successfully remove the poison from models pre-trained using the strong objective, indicating that having distinct pre-training and cleaning objectives does not ensure removal of poison. }
    \label{fig:deep-clustering-and-cheating}
\end{figure}

In the latter case, when we use clustering-based pseudo-label assignment for every image in the cleaning dataset, we learn to predict the assigned pseudo-label ($\hat{y}$) with the help of a linear classifier on top of the vision encoder using cross-entropy loss ($ \mathcal{L}_{\text{Xent}}$). Let $W \in \mathbb{R}^{d \times 1000}$ be the linear classifier mapping visual features ($\mathbb{R}^d$) to one of the 1000 pseudo-labels. For a given datapoint $(I_j, T_j)$ with assigned pseudo-label $\hat{y}_j$, deep clustering's objective becomes
\begin{equation}
    \mathcal{L}_{\text{DeepClust}}^{ft}(f_I, W; I_j, \hat{y}_j) = \mathcal{L}_{\text{Xent}}(W^Tf_I(I_j);\hat{y}_j),
\end{equation}
and therefore overall objective becomes $\mathcal{L}^{ft}_{\text{MMCL}} + \mathcal{L}^{ft}_{\text{SSL}} +  \mathcal{L}_{\text{DeepClust}}^{ft}$. Following \citet{deep-clustering-paper}, we re-initialize classifier head $W$ every time we re-compute pseudo-labels. 
We finetune the model trained using $\mathcal{L}^{pre}_{\text{MMCL}} + \mathcal{L}^{pre}_{\text{SSL}}$ on the CC6M dataset for 10 epochs using 8 different learning rates for both the clustering techniques (see \Cref{sec:hyperparam-details-appendix} for hyperparameter details) and measure the Top-1 zero-shot Imagenet accuracy and ASR at the end of each finetuning epoch. 

\Cref{fig:deep-clustering-and-cheating} shows the scatter plot of the Top-1 zero-shot Imagenet validation set accuracy and the ASR for this experiment. We observe that adding deep clustering does not successfully remove the poison from the model. 
This experiment shows that the ineffectiveness of CleanCLIP in cleaning the model trained with the stronger objective is not just due to the same pre-training and cleaning objectives but due to stronger poison induction. Had the difference between the pre-training and finetuning objectives been the reason for CleanCLIP's success, then this DeepClustering experiment could have successfully cleaned the model. 

\paragraph{Poison Removal using Heavy Regularization}
In this section, we attempt to remove the poison induced using $\mathcal{L}^{pre}_{\text{MMCL}} + \mathcal{L}^{pre}_{\text{SSL}}$ with heavy regularization. 
\citet{heavy-regularization} proposed heavy regularization as an approach to remove backdoors from vision and language models. To evaluate the efficacy of this approach, we finetune these models with 9 different regularization weights for 10 epochs using 8 different learning rates for each regularization weight on 100K image-text pairs (see \Cref{sec:hyperparam-details-appendix} for hyperparameter details). The regularization loss is added to three different loss objectives during the finetuning process: a) $\mathcal{L}^{ft}_{\text{MMCL}}, b) \mathcal{L}^{ft}_{\text{SSL}}$, and c) $\mathcal{L}^{ft}_{\text{MMCL}} + \mathcal{L}^{ft}_{\text{SSL}}$. 

\begin{figure}[h]
    \centering
    \begin{minipage}[t]{0.48\textwidth}
        \centering
        \begin{tikzpicture}
            \node[anchor=south west,inner sep=0] (image) at (0,0) {
                \includegraphics[width=0.9\linewidth]{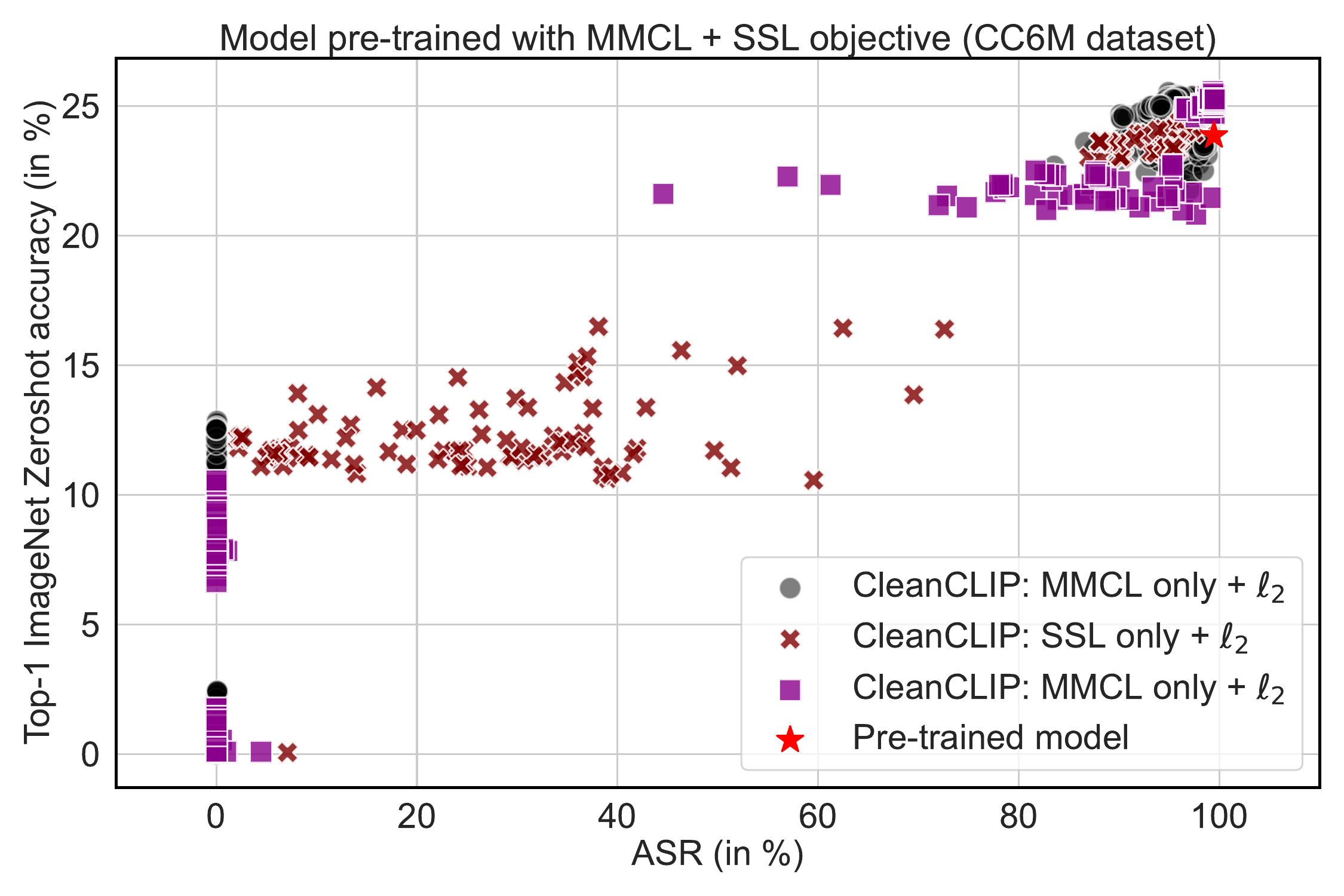}
            };
            \begin{scope}[x={(image.south east)},y={(image.north west)}]
                \draw[red, thick] (0.16,0.85) ellipse (0.06*0.66 and 0.06*1);
                \draw[black, thick] (0.9,0.85) ellipse (0.04*0.66 and 0.04*1);
                \draw[-{Stealth[length=2mm,width=2mm]}, red, thick] (0.08,0.96) -- (0.132,0.893);
                \draw[-{Stealth[length=2mm,width=2mm]}, black, thick] (0.97,0.95) -- (0.92,0.87);
            \end{scope}
        \end{tikzpicture}
        \caption{Top-1 zero-shot Imagenet validation set accuracy vs. the ASR during the cleaning process. \textbf{Takeaway:} Using regularization with any hyperparameter combinations is unable to remove the poison from models trained using $\mathcal{L}^{pre}_{\text{MMCL}} + \mathcal{L}^{pre}_{\text{SSL}}$. }
        \label{fig:clean-with-heavy-regularization}
    \end{minipage}\hfill
    \begin{minipage}[t]{0.48\textwidth}
        \centering
        \begin{tikzpicture}
            \node[anchor=south west,inner sep=0] (image) at (0,0) {
                \includegraphics[width=0.9\linewidth]{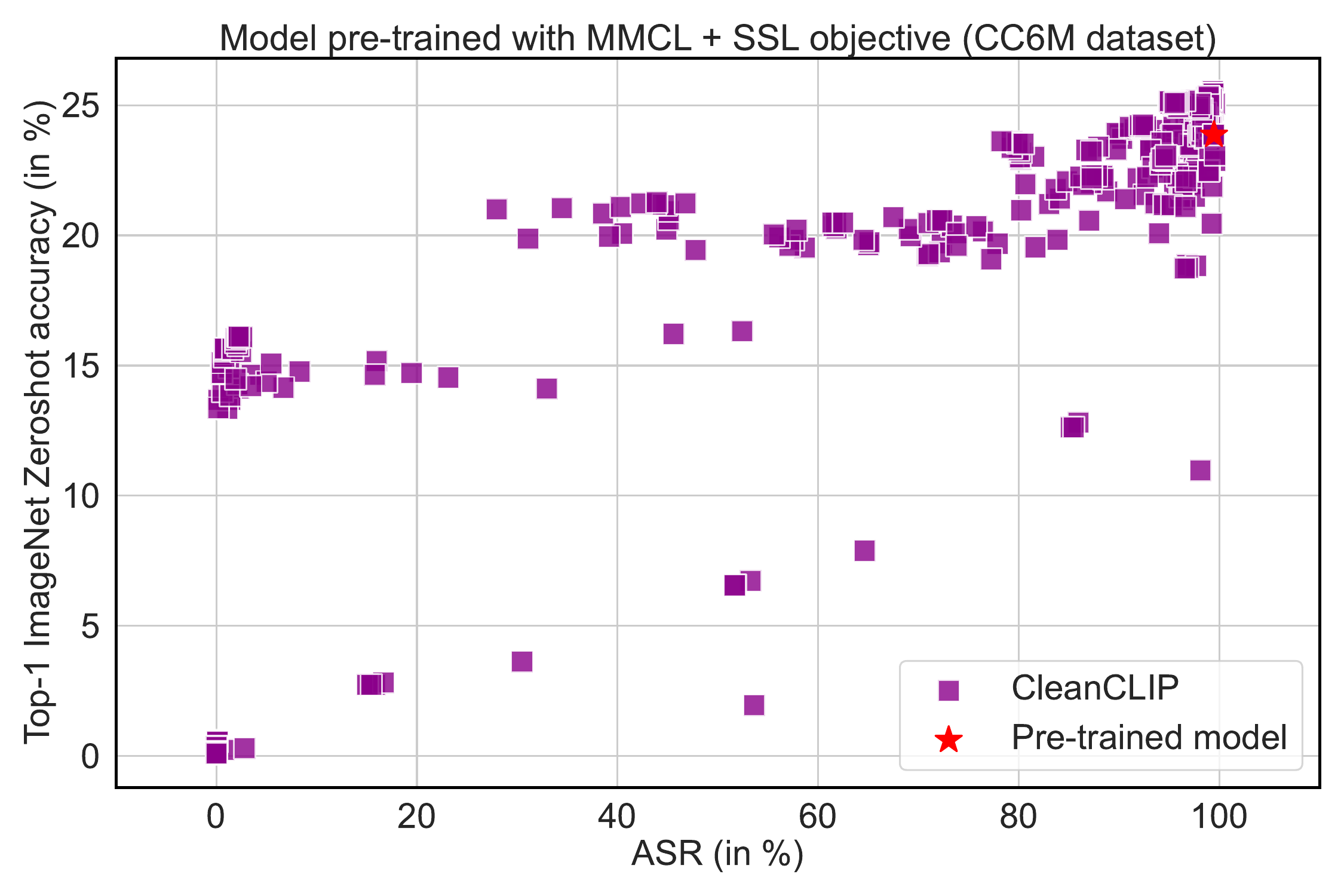}
            };
            \begin{scope}[x={(image.south east)},y={(image.north west)}]
                \draw[red, thick] (0.16,0.85) ellipse (0.06*0.66 and 0.06*1);
                \draw[black, thick] (0.9,0.85) ellipse (0.04*0.66 and 0.04*1);
                \draw[-{Stealth[length=2mm,width=2mm]}, red, thick] (0.08,0.96) -- (0.132,0.893);
                \draw[-{Stealth[length=2mm,width=2mm]}, black, thick] (0.97,0.95) -- (0.92,0.87);
            \end{scope}
        \end{tikzpicture}
        \caption{Top-1 zero-shot Imagenet validation set accuracy vs. the ASR during the cleaning process post shrinking and perturbing the weights. \textbf{Takeaway:} The shrink and perturb technique is unable to remove poison from the model that is trained using $\mathcal{L}^{pre}_{\text{MMCL}} + \mathcal{L}^{pre}_{\text{SSL}}$. }
        \label{fig:clean-with-shrink-and-perturb}
    \end{minipage}
\end{figure}

\Cref{fig:clean-with-heavy-regularization} shows the scatter plot of the Top-1 zero-shot Imagenet validation set accuracy and the ASR at the end of each finetuning epoch. We observe that no hyperparameter combination can successfully remove the poison, which shows that simply adding regularization cannot remove the strongly induced poison.

\paragraph{Poison Removal using Shrink and Perturb}
In this section, we attempt to remove the poison using the shrink and perturb technique. 
\citet{shrink-and-perturb} proposed shrink and perturb technique to improve the accuracy of a model when finetuned for a task. CleanCLIP also cleans a model by finetuning, and therefore, it could benefit from this technique. In this technique, a small noise is added to the model weights before starting finetuning, $\theta_0 \leftarrow \lambda \theta_0 + p_0 \text {, where } p \sim \mathcal{N}\left(0, \sigma^2\right) \text { and } \lambda \in (0,1)$. 
To evaluate whether this approach can help in removing the poison from a model trained using $\mathcal{L}^{pre}_{\text{MMCL}} + \mathcal{L}^{pre}_{\text{SSL}}$, we add a small noise to all its weights.  While the choice of noise scale ($\sigma$) and shrinking parameter ($\lambda$) is a hyperparameter, we experiment with 5 values of $\sigma$ for 15 values of $\lambda$. 
To determine which model to clean amongst the 75 possible models obtained from shrink and perturb described above, we measure the Top-1 zero-shot Imagenet validation set accuracy and the ASR of the models with noised weights, and selected model with the highest accuracy whose ASR was lower than 15\% for cleaning. 
We clean the selected model by finetuning it using 8 learning rates for 20 epochs on 100K image-text pairs. \Cref{fig:clean-with-shrink-and-perturb} shows the accuracy and ASR at the end of each finetuning epoch. We observe that even this approach is unable to successfully remove the poison from the models.

\paragraph{Ablation on Hyperparameters}
We also perform ablation experiments to see the impact of hyperparameters on the ability of CleanCLIP to remove poison from the $\mathcal{L}^{pre}_{\text{MMCL}} + \mathcal{L}^{pre}_{\text{SSL}}$ pre-trained model. In particular, we study the impact of the size of the cleaning dataset, the number of finetuning epochs, and the weightage of $\mathcal{L}^{ft}_{\text{SSL}}$. 
\Cref{sec:clean-larger-200k} reports the metrics when the finetuning is done on a dataset that is twice the size of the cleaning dataset used in the previous section.
\Cref{sec:finetune-longer-epochs} reports the metrics when the finetuning is done for up to 100 epochs, i.e., 5$\times$ longer than in the previous section. 
\Cref{sec:higher-ssl-lambda} reports the accuracy and ASR metrics when the finetuning is done with higher weight to the $\mathcal{L}^{ft}_{\text{SSL}}$ loss. 
While changing the hyperparameters helps improve the Pareto-frontier curve, none of the three ablations can successfully remove the strongly induced poison.

\paragraph{Takeaways} These experiments demonstrate that making the pre-training and finetuning objectives distinct from each other and perturbation techniques like heavy regularization \citep{heavy-regularization} and shrink-and-perturb \citep{shrink-and-perturb} are not enough to remove the strongly induced poison. Moreover, we observe that while doing a more fine-grained search on different hyperparameters helps slightly improve the Pareto-frontier curve, none of the hyperparameter ablations can successfully remove the poison. Since removing the poison from such a model is an open research problem, a practitioner who is engaged in training models using web-curated data should consider training their model with only $\mathcal{L}^{pre}_{\text{MMCL}}$ so that they can clean their model on a carefully curated image-text pair dataset to remove potential backdoors from it, even when there are a few poisoned examples in the finetuning dataset.

\section{Conclusions}
\label{sec:conclusions}

Through our extensive hyperparameter search and ablation experiments, we unveil a critical limitation of the current state-of-the-art poison mitigation technique for multimodal models, CleanCLIP. It fails to effectively remove backdoor poisoning when a model is trained using stronger objectives like the combination of multimodal contrastive learning (MMCL) and intramodal self-supervised learning (SSL). This objective is common in popular approaches like SLIP \citep{mu2022slip} and has demonstrated superior accuracy over training with only the MMCL objective. 
Our experiments show that this vulnerability persists irrespective of the scale of the pre-training and the cleaning datasets, irrespective of the manner of poison induction (from scratch or by finetuning), and irrespective of the specific backdoor attack. 

Particularly concerning is the unstable cleaning trajectory in models trained using the stronger objective (\Cref{fig:mmcl-ssl-finetuning-trajectory}). Often unaware of the specific backdoor attack, practitioners face challenges in determining the optimal point to halt the cleaning process. This instability can lead to suboptimal models, as continued finetuning can decrease accuracy and increase attack success rate (ASR). Furthermore, our findings highlight the critical assumption of a completely poison-free cleaning dataset for CleanCLIP's effectiveness, an assumption that is rarely met in practical scenarios. This becomes particularly problematic with the use of stronger pre-training objectives.
The models trained with the simpler MMCL objective evade both these issues by having stable cleaning trajectories and amenability to poison reduction even under non-ideal conditions. 

Given these insights, we urge practitioners to consider training their models using the simpler MMCL objective. Even though this might slightly hurt the accuracy, it significantly enhances its amenability to remove backdoors. Our recommendation would also circumvent the issue of knowing when to halt the cleaning procedure, as more finetuning epochs would not hurt the model's accuracy and ASR. Further, it will also be beneficial when cleaning data is not entirely poison-free.
Our work underscores the formidable challenge of defending models against backdoor attacks, an open research problem. We encourage future defense methods to robustly test their approach against various pre-training objectives and we invite the community to develop robust defense methods. 


\bibliographystyle{tmlr}
\bibliography{refs}

\clearpage

\appendix
\begin{table*}[t]
    \centering
    \caption{This table shows the original Top-1 zero-shot Imagenet validation set accuracy and ASR for the models trained with $\mathcal{L}^{pre}_{\text{MMCL}}$ and $\mathcal{L}^{pre}_{\text{MMCL}} + \mathcal{L}^{pre}_{\text{SSL}}$ on the CC3M and CC6M datasets. These models are the input to CleanCLIP approach to remove their poison. \textbf{Takeaway:} The models trained with $\mathcal{L}^{pre}_{\text{MMCL}} + \mathcal{L}^{pre}_{\text{SSL}}$ achieve higher accuracy than their counterparts trained with $\mathcal{L}^{pre}_{\text{MMCL}}$ solely (except when poisoned by finetuning). Adding BadNet poison to a mere 0.05\% of the training data achieves almost a 100\% ASR when trained from scratch and almost 90\% ASR when induced by finetuning.  }
    \small
    
    \begin{tabular}{c c c c c c c}
        \toprule
    Poison Induction & Backdoor & Objective & \multicolumn{4}{c}{Dataset} \\ \cmidrule(lr){4-7}
    &  &      & \multicolumn{2}{c}{CC3M}   & \multicolumn{2}{c}{CC6M}  \\ \cmidrule(lr){4-5} \cmidrule(lr){6-7}
    &  &      & Accuracy ($\uparrow$)    & ASR ($\downarrow$)   & Accuracy ($\uparrow$) & ASR ($\downarrow$) \\ \midrule
    
    From Scratch & BadNet &    $\mathcal{L}^{pre}_{\text{MMCL}}$              &   16.00\%   &  99.88\%     &    23.76\%    & 99.98\%    \\
    From Scratch & BadNet &    $\mathcal{L}^{pre}_{\text{MMCL}} + \mathcal{L}^{pre}_{\text{SSL}}$        &   17.04\%   &  99.03\%     &    23.86\%    & 99.45\%     \\ \\
    
    From Scratch & Label-Consistent &    $\mathcal{L}^{pre}_{\text{MMCL}}$       &   --   &  --     &  22.96\%    &  88.27\%       \\

    From Scratch & Label-Consistent &  $\mathcal{L}^{pre}_{\text{MMCL}} + \mathcal{L}^{pre}_{\text{SSL}}$        &   --   &  --     &  23.09\%    &  88.01\%        \\ \\

    Finetuning from Ckpt & BadNet & $\mathcal{L}^{pre}_{\text{MMCL}}$       &   --   &  --     & 60.30\%    & 99.20\%   \\

    Finetuning from Ckpt & BadNet & $\mathcal{L}^{pre}_{\text{MMCL}} + \mathcal{L}^{pre}_{\text{SSL}}$        &   --   &  --     &    59.81\%    & 90.24\%  \\ 
    \bottomrule
    \end{tabular}
    \label{tab:starting-accuracy-asr}
\vspace{1em}

\end{table*}

\section{Pre-Training Details}
\label{sec:pre-training-details-and-starting-acc-asr}
We train all the models on 8 Nvidia A100 GPUs for 64 epochs. We use an initial learning rate of 0.001 for the models trained from scratch, and for the models where poison is induced by finetuning from a pre-trained checkpoint, we use an initial learning rate of $5e-7$. We use cosine scheduling and 10000 warmup steps with AdamW optimizer \citep{Loshchilov2017AdamW} for training. The model trained with $\mathcal{L}^{pre}_{\text{MMCL}}$ uses a batch size of 256, whereas the model trained with the $\mathcal{L}^{pre}_{\text{MMCL}} + \mathcal{L}^{pre}_{\text{SSL}}$ uses a batch size of 128. 

\begin{figure*}[t]
    \centering
    \includegraphics[width=0.8\textwidth]{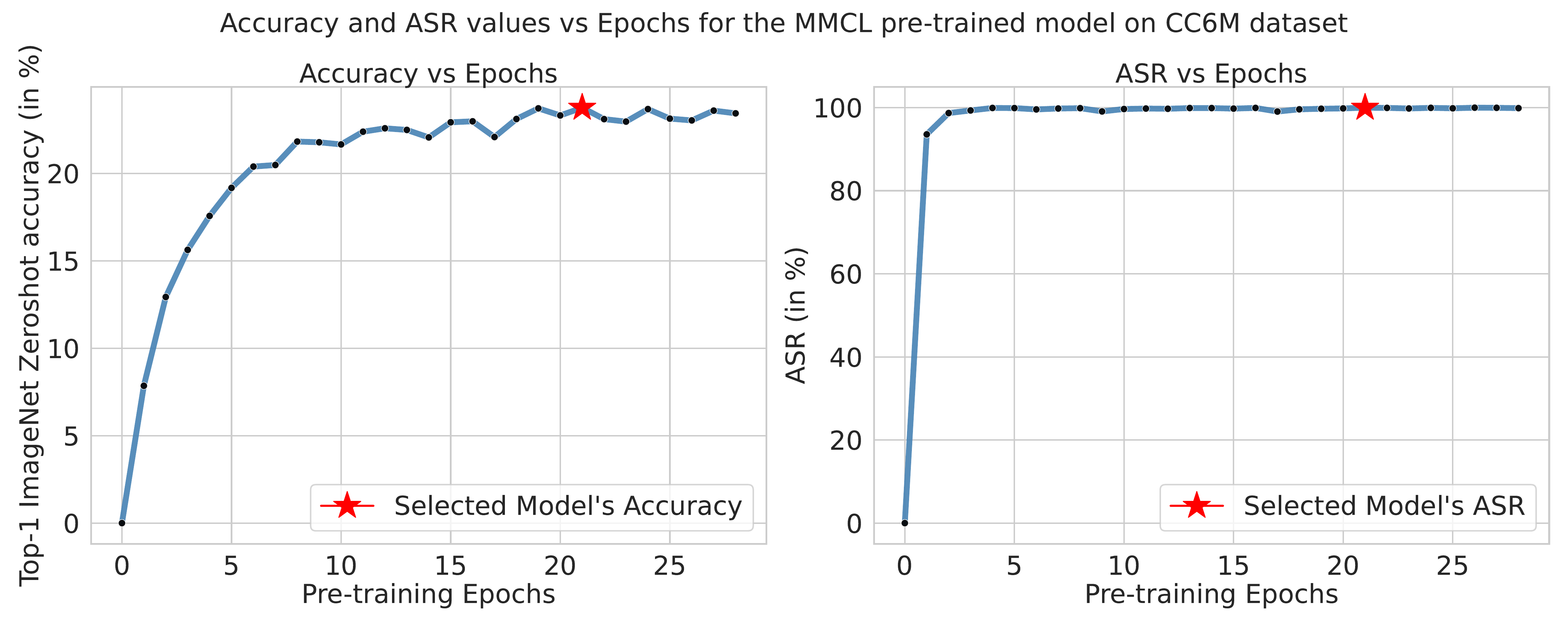}
    \caption{Top-1 zero-shot Imagenet validation set accuracy and the ASR of the model during its pre-training on the CC6M dataset using $\mathcal{L}^{pre}_{\text{MMCL}}$. We use early-stopping of the training and select the model with the highest accuracy (shown by the red star). It has 23.76\% accuracy and 99.98\% ASR.}
    \label{fig:accuracy_asr_pretraining_cc6m_using_mmcl}
\end{figure*}

\begin{figure*}[t]
    \centering
    \includegraphics[width=0.8\textwidth]{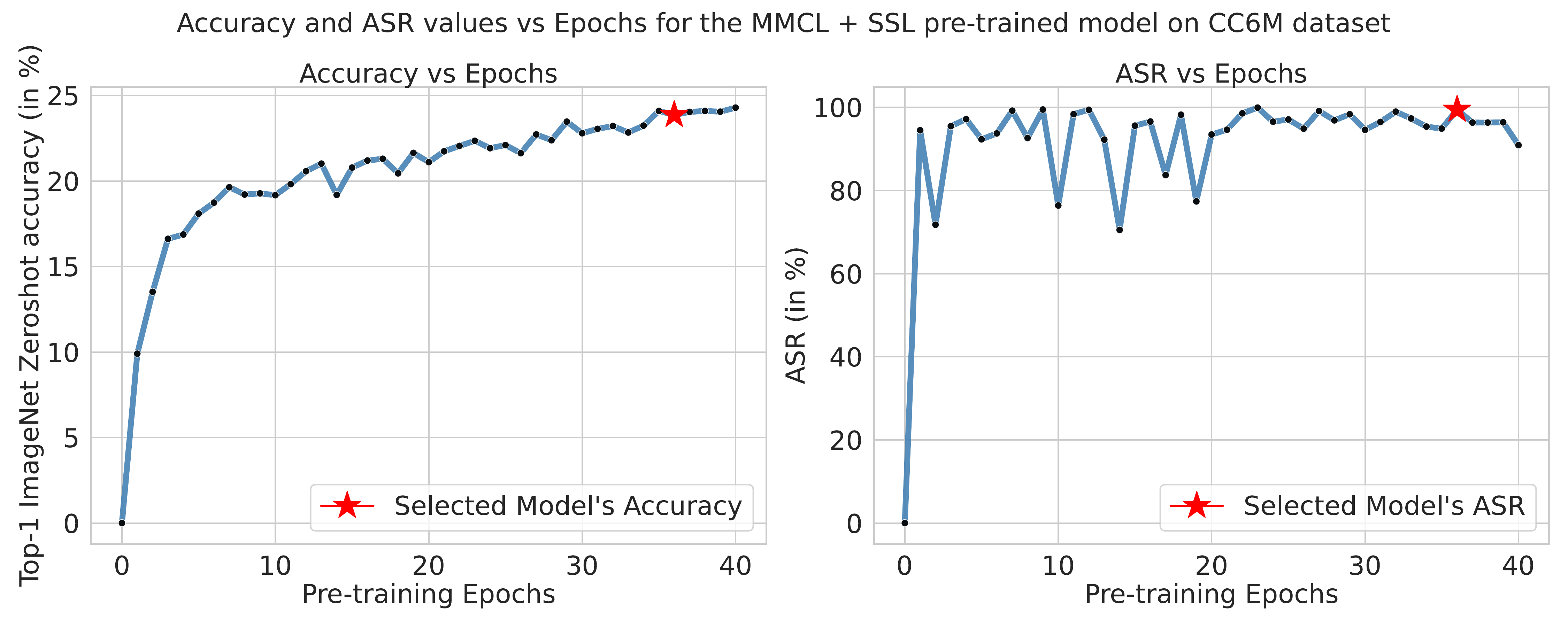}
    \caption{Top-1 zero-shot Imagenet validation set accuracy and the ASR of the model during its pre-training on the CC6M dataset using $\mathcal{L}^{pre}_{\text{MMCL}} + \mathcal{L}^{pre}_{\text{SSL}}$. We use early-stopping of the training and select the model with the accuracy closest to the $\mathcal{L}^{pre}_{\text{MMCL}}$ pre-trained model's accuracy (shown by the red star). The selected model has 23.86\% accuracy and 99.45\% ASR.}
    \label{fig:accuracy_asr_pretraining_cc6m_using_mmcl_ssl}
\end{figure*}

We show the change in accuracy and ASR with the training epochs for the model trained from scratch with BadNet attack using $\mathcal{L}^{pre}_{\text{MMCL}}$ in \Cref{fig:accuracy_asr_pretraining_cc6m_using_mmcl} and using $\mathcal{L}^{pre}_{\text{MMCL}} + \mathcal{L}^{pre}_{\text{SSL}}$ in \Cref{fig:accuracy_asr_pretraining_cc6m_using_mmcl_ssl}. We use early-stopping for the model trained with $\mathcal{L}^{pre}_{\text{MMCL}}$ and choose the model with the highest accuracy. For $\mathcal{L}^{pre}_{\text{MMCL}} + \mathcal{L}^{pre}_{\text{SSL}}$ pre-training, we choose the model that has the closest accuracy to the $\mathcal{L}^{pre}_{\text{MMCL}}$ trained model. 
\Cref{tab:starting-accuracy-asr} shows the accuracy and the ASR for all the models we select in this paper for poison removal. 

\section{Performance of the Pre-trained Models}
\Cref{tab:starting-accuracy-asr} shows the Top-1 zero-shot Imagenet Validation set accuracy and the ASR of models that were selected for both the pre-training objectives. 

\section{Templates for Text-Embedding Computation}
\label{sec:appendix-text-template}

\begin{lstlisting}
`a bad photo of a {class}.', `a photo of many {class}.', `a sculpture of a {class}.', `a photo of the hard to see {class}.', `a low resolution photo of the {class}.', `a rendering of a {class}.', `graffiti of a {class}.', `a bad photo of the {class}.', `a cropped photo of the {class}.', `a tattoo of a {class}.', `the embroidered {class}.', `a photo of a hard to see {class}.', `a bright photo of a {class}.', `a photo of a clean {class}.', `a photo of a dirty {class}.', `a dark photo of the {class}.', `a drawing of a {class}.', `a photo of my {class}.', `the plastic {class}.', `a photo of the cool {class}.', `a close-up photo of a {class}.', `a black and white photo of the {class}.', `a painting of the {class}.', `a painting of a {class}.', `a pixelated photo of the {class}.', `a sculpture of the {class}.', `a bright photo of the {class}.', `a cropped photo of a {class}.', `a plastic {class}.', `a photo of the dirty {class}.', `a jpeg corrupted photo of a {class}.', `a blurry photo of the {class}.', `a photo of the {class}.', `a good photo of the {class}.', `a rendering of the {class}.', `a {class} in a video game.', `a photo of one {class}.', `a doodle of a {class}.', `a close-up photo of the {class}.', `a photo of a {class}.', `the origami {class}.', `the {class} in a video game.', `a sketch of a {class}.', `a doodle of the {class}.', `a origami {class}.', `a low resolution photo of a {class}.', `the toy {class}.', `a rendition of the {class}.', `a photo of the clean {class}.', `a photo of a large {class}.', `a rendition of a {class}.', `a photo of a nice {class}.', `a photo of a weird {class}.', `a blurry photo of a {class}.', `a cartoon {class}.', `art of a {class}.', `a sketch of the {class}.', `a embroidered {class}.', `a pixelated photo of a {class}.', `itap of the {class}.', `a jpeg corrupted photo of the {class}.', `a good photo of a {class}.', `a plushie {class}.', `a photo of the nice {class}.', `a photo of the small {class}.', `a photo of the weird {class}.', `the cartoon {class}.', `art of the {class}.', `a drawing of the {class}.', `a photo of the large {class}.', `a black and white photo of a {class}.', `the plushie {class}.', `a dark photo of a {class}.', `itap of a {class}.', `graffiti of the {class}.', `a toy {class}.', `itap of my {class}.', `a photo of a cool {class}.', `a photo of a small {class}.', `a tattoo of the {class}.'
\end{lstlisting}

\section{Hyperparameter Details}
\label{sec:hyperparam-details-appendix}

In this section we provide details of the hyperparameters we used for the cleaning experiments.

\subsection{Cleaning of the Model Pre-trained on CC3M dataset using MMCL}

Cleaning Epochs: 20 \newline
Learning rates (13 values):
\{1e-5, 4e-5, 8e-5, 1e-4, 1.5e-4, 2e-4, 2.5e-4, 3e-4, 3.5e-4, 4e-4, 8e-4, 1e-3, 4e-3\}\newline
MMCL weight: 1 \newline
SSL weight: 1 \newline
Size of the Cleaning Dataset: 1,00,000

\subsection{Cleaning of the Model Pre-trained on CC3M dataset using MMCL and SSL}

Cleaning Epochs: 20 \newline
Learning rates (13 values):
\{1e-5, 4e-5, 8e-5, 1e-4, 1.5e-4, 2e-4, 2.5e-4, 3e-4, 3.5e-4, 4e-4, 8e-4, 1e-3, 4e-3\} \newline
MMCL weight: 1 \newline
SSL weight: 1 \newline
Size of the Cleaning Dataset: 1,00,000

\subsection{Cleaning of the Model Pre-trained on CC6M dataset using MMCL}

Cleaning Epochs: 20 \newline
Learning rates (12 values): \{1e-9, 5e-9, 1e-8, 5e-8, 1e-7, 3e-7, 7e-7, 1e-6, 3e-6, 7e-6, 1e-5, 3e-5\}  \newline
MMCL weight: 1 \newline
SSL weight: 1 \newline
Size of the Cleaning Dataset: 1,00,000

\subsection{Cleaning of the Model Pre-trained on CC6M dataset using MMCL and SSL}

Cleaning Epochs: 20 \newline
Learning rates (23 values): 
\{1e-9, 5e-9, 1e-8, 5e-8, 1e-7, 3e-7, 7e-7, 1e-6, 3e-6, 7e-6, 1e-5, 4e-5, 5e-5, 6e-5, 7e-5, 9e-5, 1e-4, 3e-4, 4e-4, 5e-4, 6e-4, 1e-3, 3e-3\} \newline
MMCL weight: 1 \newline
SSL weight: 1 \newline
Size of the Cleaning Dataset: 1,00,000

\subsection{Cleaning of the Model where Poison is induced via Finetuning with MMCL on CC6M dataset}

Cleaning Epochs: 20 \newline
Learning rates (69 values): \{5e-05, 4.9e-05, 4.8e-05, 4.7e-05, 4.6e-05, 4.5e-05, 4.4e-05, 4.3e-05, 4.25e-05, 4.2e-05, 4.1e-05, 4e-05, 3.9e-05, 3.8e-05, 3.75e-05, 3.7e-05, 3.6e-05, 3.5e-05, 3.4e-05, 3.3e-05, 3.2e-05, 3.1e-05, 3e-05, 2.9e-05, 2.8e-05, 2.7e-05, 2.6e-05, 2.5e-05, 2.4e-05, 2.3e-05, 2.2e-05, 2.1e-05, 2e-05, 1.9e-05, 1.8e-05, 1.7e-05, 1.6e-05, 1.5e-05, 1.4e-05, 1.3e-05, 1.2e-05, 1.1e-05, 1e-05, 9e-06, 8e-06, 7e-06, 6e-06, 5e-06, 4.9e-06, 4.75e-06, 4.5e-06, 4.25e-06, 4e-06, 3.75e-06, 3.5e-06, 3.25e-06, 3e-06, 2.75e-06, 2.5e-06, 2.25e-06, 2e-06, 1.75e-06, 1.5e-06, 1.25e-06, 1e-06, 5e-07, 1e-07, 5e-08, 1e-08\} \newline
MMCL weight: 1 \newline
SSL weight: 1 \newline
Size of the Cleaning Dataset: 1,00,000

\subsection{Cleaning of the Model where Poison is induced via Finetuning with MMCL + SSL on CC6M dataset}

Cleaning Epochs: 20 \newline
Learning rates (85 values): \{0.005, 0.001, 0.0005, 0.0001, 5e-05, 5e-05, 4.9e-05, 4.8e-05, 4.75e-05, 4.7e-05, 4.6e-05, 4.5e-05, 4.5e-05, 4.4e-05, 4.3e-05, 4.25e-05, 4.2e-05, 4.1e-05, 4e-05, 4e-05, 3.9e-05, 3.8e-05, 3.75e-05, 3.7e-05, 3.6e-05, 3.5e-05, 3.5e-05, 3.4e-05, 3.3e-05, 3.25e-05, 3.2e-05, 3.1e-05, 3e-05, 3e-05, 2.9e-05, 2.8e-05, 2.75e-05, 2.7e-05, 2.6e-05, 2.5e-05, 2.5e-05, 2.4e-05, 2.3e-05, 2.25e-05, 2.2e-05, 2.1e-05, 2e-05, 2e-05, 1.9e-05, 1.8e-05, 1.75e-05, 1.7e-05, 1.6e-05, 1.5e-05, 1.5e-05, 1.4e-05, 1.4e-05, 1.3e-05, 1.3e-05, 1.2e-05, 1.2e-05, 1.1e-05, 1.1e-05, 1e-05, 1e-05, 9e-06, 9e-06, 8e-06, 8e-06, 7e-06, 7e-06, 6e-06, 6e-06, 5e-06, 5e-06, 1e-06, 1e-06, 5e-07, 5e-07, 1e-07, 1e-07, 5e-08, 5e-08, 1e-08, 1e-08\} \newline
MMCL weight: 1 \newline
SSL weight: 1 \newline
Size of the Cleaning Dataset: 1,00,000

\subsection{Cleaning of the Model with Label Consistent Poisoning trained with MMCL on CC6M dataset}

Cleaning Epochs: 20  \newline
Learning rates (44 values): \{5e-05, 4.75e-05, 4.25e-05, 4e-05, 3.8e-05, 3.75e-05, 3.7e-05, 3.6e-05, 3.5e-05, 3.4e-05, 3.3e-05, 3.2e-05, 3.1e-05, 3e-05, 2.9e-05, 2.8e-05, 2.7e-05, 2.6e-05, 2.5e-05, 2.4e-05, 2.3e-05, 2.2e-05, 2.1e-05, 2e-05, 1.9e-05, 1.8e-05, 1.7e-05, 1.6e-05, 1.5e-05, 1.4e-05, 1.3e-05, 1.2e-05, 1.1e-05, 1e-05, 9e-06, 8e-06, 7e-06, 6e-06, 5e-06, 1e-06, 5e-07, 1e-07, 5e-08, 1e-08\} \newline
MMCL weight: 1  \newline
SSL weight: 1  \newline
Size of the Cleaning Dataset: 1,00,000

\subsection{Cleaning of the Model with Label Consistent Poisoning trained with MMCL + SSL on CC6M dataset}

Cleaning Epochs: 20  \newline
Learning rates (71 values): \{0.007, 0.006, 0.005, 0.004, 0.003, 0.002, 0.001, 0.0009, 0.0008, 0.0007, 0.0006, 0.0005, 0.0004, 0.0003, 0.0002, 0.00018, 0.00017, 0.00016, 0.00014, 0.00013, 0.00012, 0.00011, 0.0001, 9e-05, 8e-05, 7e-05, 6e-05, 5e-05, 4.75e-05, 4.25e-05, 4e-05, 3.8e-05, 3.75e-05, 3.7e-05, 3.6e-05, 3.5e-05, 3.4e-05, 3.3e-05, 3.2e-05, 3.1e-05, 3e-05, 2.9e-05, 2.8e-05, 2.7e-05, 2.6e-05, 2.5e-05, 2.4e-05, 2.3e-05, 2.2e-05, 2.1e-05, 2e-05, 1.9e-05, 1.8e-05, 1.7e-05, 1.6e-05, 1.5e-05, 1.4e-05, 1.3e-05, 1.2e-05, 1.1e-05, 1e-05, 9e-06, 8e-06, 7e-06, 6e-06, 5e-06, 1e-06, 5e-07, 1e-07, 5e-08, 1e-08\} \newline
MMCL weight: 1  \newline
SSL weight: 1  \newline
Size of the Cleaning Dataset: 1,00,000

\subsection{Cleaning of the Model Pre-trained on the CC3M dataset under Non-Ideal Conditions}

Cleaning Epochs: 20  \newline
Learning rates (8 values): \{1e-7, 3e-7, 7e-7, 1e-6, 3e-6, 7e-6, 1e-5, 3e-5\} \newline
MMCL weight: 1  \newline
SSL weight: 1  \newline
Size of the Cleaning Dataset: 1,00,000

\subsection{Cleaning of the Model Pre-trained on the CC6M dataset under Non-Ideal Conditions}

Cleaning Epochs: 20  \newline
Learning rates (19 values): \{1e-8, 5e-8, 1e-7, 3e-7, 5e-7, 7e-7, 1e-6, 3e-6, 5e-6, 7e-6, 1e-5, 3e-5, 5e-5, 1e-4, 3e-4, 5e-4, 7e-4, 1e-4, 3e-4\} 
\newline
MMCL weight: 1  \newline
SSL weight: 1 \newline
Size of the Cleaning Dataset: 1,00,000

\subsection{Cleaning of the Model Pre-trained on the CC6M dataset using an Objective distinct from Pre-training}

Cleaning Epochs: 20  \newline
Learning rates (9 values): \{5e-6, 1e-5, 5e-5, 1e-4, 2e-4, 3e-4, 4e-4, 5e-4, 1e-3\} \newline
MMCL weight: 1  \newline
SSL weight: 1  \newline
Deep Clustering Loss Weight (8 values): \{0.1, 0.5, 1, 2, 5, 10, 20, 50\} \newline
Size of the Cleaning Dataset: 1,00,000

\subsection{Cleaning of the Model Pre-trained on the CC6M dataset using Heavy Regularization}

Cleaning Epochs: 20  \newline
Learning rates (8 values): \{3e-6, 7e-6, 1e-5, 3e-5, 1e-4, 5e-4, 1e-3, 5e-3\} \newline
MMCL weight: 1  \newline
SSL weight: 1  \newline
$\ell_2$ weight (9 values): \{0.2, 0.5, 1, 2, 5, 10, 20, 50, 100\} \newline
Size of the Cleaning Dataset: 1,00,000

\subsection{Cleaning of the Model Pre-trained on the CC6M dataset using Shrink and Perturb}

Cleaning Epochs: 20  \newline
Learning rates (9 values): \{1e-5, 2e-5, 4e-5, 5e-5, 7e-5, 9e-5, 1e-4, 2e-4, 1e-3\} \newline
MMCL weight: 1  \newline
SSL weight: 1  \newline
Shrink $\lambda$ (17 values): \{0.1, 0.2, 0.3, 0.4, 0.5, 0.6, 0.7, 0.8, 0.9, 0.92, 0.93, 0.95, 0.96, 0.97, 0.98, 0.99, 1\} \newline
Perturb p (15 values): \{1e-5, 1e-4, 1e-3, 0.01, 0.02, 0.04, 0.06, 0.08, 0.1, 0.4, 0.8, 1, 2, 3, 4\} \newline
Size of the Cleaning Dataset: 1,00,000

\subsection{Cleaning of the Model Pre-trained on the CC6M dataset using a Larger Cleaning Dataset}

Cleaning Epochs: 20  \newline
Learning rates (14 values): \{1e-9, 5e-9, 1e-8, 5e-8, 1e-7, 3e-7, 7e-7, 1e-6, 3e-6, 7e-6, 1e-5, 3e-5, 1e-4, 1e-3\}
\newline
MMCL weight: 1  \newline
SSL weight: 1 \newline
Size of the Cleaning Dataset: 2,00,000

\subsection{Cleaning of the Model Pre-trained on the CC6M dataset with More Finetuning Epochs}

\textbf{Cleaning Epochs (2 values):} 50, 100  \newline
Learning rates (13 values): \{1e-5, 2e-5, 3e-5, 4e-5, 5e-5, 6e-5, 7e-5, 9e-5, 1e-4, 2e-4, 3e-4, 4e-4, 5e-4\}
\newline
MMCL weight: 1  \newline
SSL weight: 1 \newline
Size of the Cleaning Dataset: 1,00,000

\subsection{Cleaning of the Model Pre-trained on the CC6M dataset using a Larger Weights for SSL Term}

Cleaning Epochs: 20  \newline
Learning rates (4 values): \{5e-5, 1e-4, 5e-4, 1e-3\} \newline
MMCL weight: 1  \newline
SSL weight (4 values): \{2, 4, 6, 8\} \newline
Size of the Cleaning Dataset: 1,00,000

\section{CC3M Results}

\subsection{Findings for the Models trained on the CC3M Dataset}
\label{sec:cc3m-main-results-appendix}

\begin{figure*}[h]
    \centering
            
    
            

    \begin{tikzpicture}
        \node[anchor=south west,inner sep=0] (image) at (0,0) {
            \includegraphics[width=\textwidth]{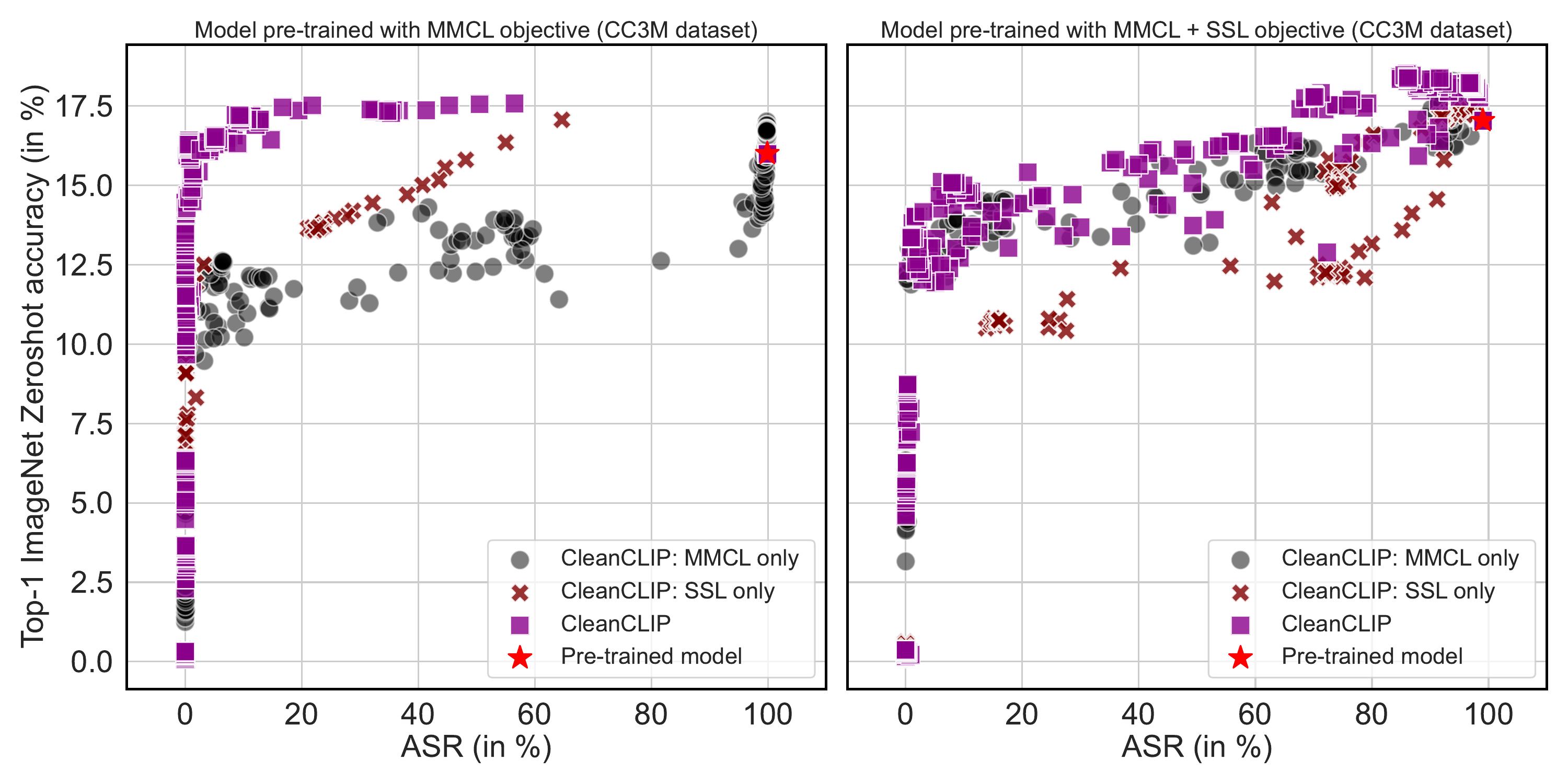}
        };
        \begin{scope}[x={(image.south east)},y={(image.north west)}]
            \draw[red, thick] (0.125,0.835) ellipse (0.06*0.5 and 0.06*1);
            \draw[red, thick] (0.58,0.855) ellipse (0.06*0.5 and 0.06*1);
            \draw[-{Stealth[length=2mm,width=2mm]}, red, thick] (0.065,0.95) -- (0.106,0.878);
            \draw[-{Stealth[length=2mm,width=2mm]}, red, thick] (0.52,0.98) -- (0.56,0.90);

            \draw[black, thick] (0.49,0.8) ellipse (0.03*0.5 and 0.03*1);
            \draw[black, thick] (0.945,0.845) ellipse (0.03*0.5 and 0.03*1);
            \draw[-{Stealth[length=2mm,width=2mm]}, black, thick] (0.54,0.9) -- (0.498,0.823);
            \draw[-{Stealth[length=2mm,width=2mm]}, black, thick] (1,0.95) -- (0.95,0.87);

            \draw[red, thick] (0.58, 0.855) -- (0.58, 0.72); 
            \draw[red, thick] (0.57, 0.855) -- (0.59, 0.855); 
            \draw[red, thick] (0.57, 0.72) -- (0.59, 0.72); 
            \node[right, text=red] at (0.59, 0.80) {20\% drop}; 
        \end{scope}
    \end{tikzpicture}

    \caption{Top-1 zero-shot Imagenet validation set accuracy vs. the ASR, measured at the end of each cleaning epoch for the models pre-trained on the CC3M dataset. The finetuning is done with each of the three losses as mentioned above. The red star in the top right corner (encircled in the black circle) corresponds to the original model's accuracy and ASR. For a successful cleaning, there should be models that maintain the original model's accuracy while having a low ASR (indicated by the red circle). \textbf{Takeaway:} CleanCLIP successfully cleans the model pre-trained with $\mathcal{L}^{pre}_{\text{MMCL}}$ (left), while it fails for the model pre-trained with $\mathcal{L}^{pre}_{\text{MMCL}} + \mathcal{L}^{pre}_{\text{SSL}}$ (right).}
    \label{fig:CC3M-cleaning-plots}
\end{figure*}

\Cref{fig:CC3M-cleaning-plots} shows the scatter plot of the Top-1 zero-shot Imagenet-1K validation set accuracy and the ASR of the models at the end of each cleaning epoch for the models pre-trained on the CC3M dataset. We observe that:
\begin{enumerate}[leftmargin=7pt,topsep=0pt,itemsep=2pt]
    \item $\mathcal{L}^{ft}_{\text{MMCL}}$ and $\mathcal{L}^{ft}_{\text{SSL}}$ individually are ineffective cleaning losses as they cause a significant drop in accuracy for lowering the ASR for both the pre-training objectives. 

    \item $\mathcal{L}^{ft}_{\text{MMCL}} + \mathcal{L}^{ft}_{\text{SSL}}$ serves as an effective cleaning loss for the model pre-trained with $\mathcal{L}^{pre}_{\text{MMCL}}$ (left plot). The cleaned models maintain the accuracy of the original model while getting a low ASR, which is successful clean. 
    However, it does not lead to an effective cleaning of the model pre-trained with $\mathcal{L}^{pre}_{\text{MMCL}} + \mathcal{L}^{pre}_{\text{SSL}}$. The models with low ASR ($\le$ 5\%) lose about 20\% of the original model's accuracy. 
\end{enumerate}

\subsection{Findings for the Models trained on the CC3M Dataset when Cleaning under Non-ideal Conditions}
\label{sec:cc3m-clean-with-slight-poison-results-appendix}

\begin{figure*}
    \centering
    \begin{tikzpicture}
        \node[anchor=south west,inner sep=0] (image) at (0,0) {
            \includegraphics[width=\textwidth]{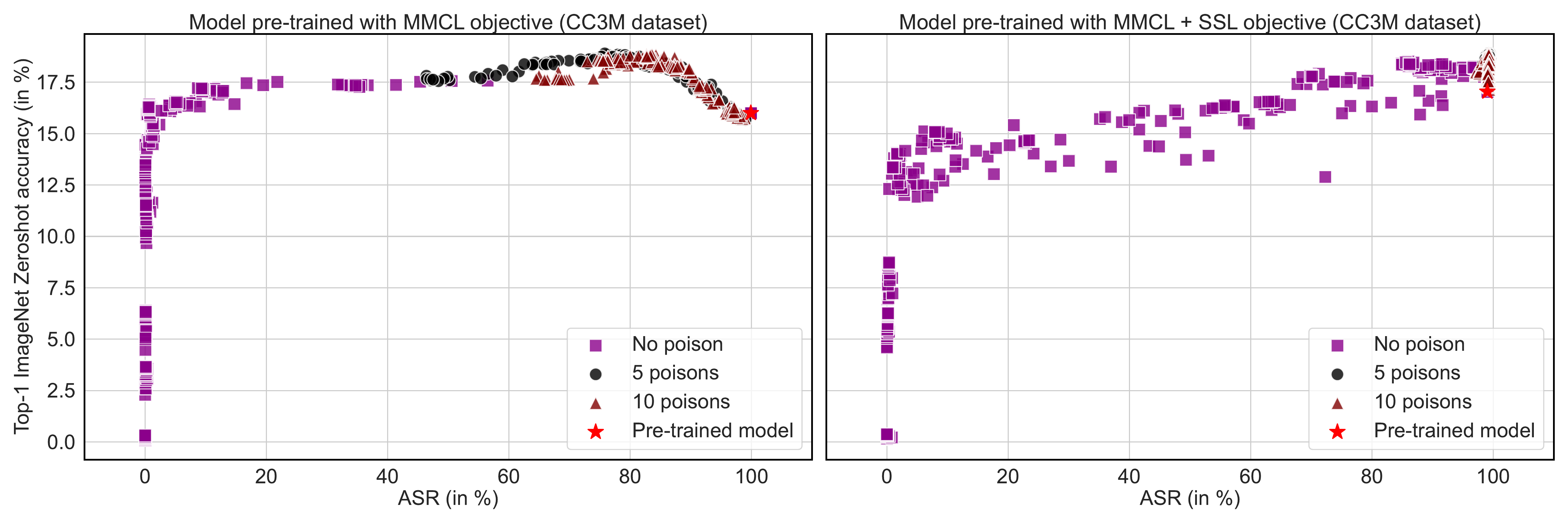}
        };
        \begin{scope}[x={(image.south east)},y={(image.north west)}]
            \draw[red, thick] (0.09,0.80) ellipse (0.06*0.33 and 0.06*1);
            \draw[red, thick] (0.57,0.84) ellipse (0.06*0.33 and 0.06*1);
    
            \draw[-{Stealth[length=2mm,width=2mm]}, red, thick] (0.043,0.907) -- (0.074,0.832);
            \draw[-{Stealth[length=2mm,width=2mm]}, red, thick] (0.525,0.955) -- (0.556,0.88);

            \draw[black, thick] (0.48,0.78) ellipse (0.04*0.33 and 0.04*1);
            \draw[black, thick] (0.95,0.83) ellipse (0.04*0.33 and 0.04*1);
            \draw[-{Stealth[length=2mm,width=2mm]}, black, thick] (0.52,0.84) -- (0.49,0.80);
            \draw[-{Stealth[length=2mm,width=2mm]}, black, thick] (1,0.95) -- (0.96,0.85);
        \end{scope}
      \end{tikzpicture}
    \caption{Scatter plot of the Top-1 zero-shot Imagenet validation set accuracy vs. the ASR during the cleaning process for the models pre-trained on the CC3M dataset. The finetuning is done with $\mathcal{L}^{ft}_{\text{MMCL}} + \mathcal{L}^{ft}_{\text{SSL}}$. We measure accuracy and ASR at the end of each epoch. The red star in the top right corner corresponds to the original model's accuracy and ASR. For a successful cleaning, there should be models that maintain the original model's accuracy while having a low ASR (indicated by the red circle). \textbf{Takeaway:} Even having 5 poisons in the cleaning dataset (i.e. 0.005\% of the dataset, which is 10$\times$ cleaner than the pre-training data) hurts the cleaning process for both pre-training objectives, and $\mathcal{L}^{pre}_{\text{MMCL}} + \mathcal{L}^{pre}_{\text{SSL}}$ pre-trained models are hurt worse. }
    \label{fig:CC3M-cleaning-plots-slight-poison}
\end{figure*}

\Cref{fig:CC3M-cleaning-plots-slight-poison} shows the scatter plot of the Top-1 zero-shot Imagenet validation set accuracy and the ASR at the end of each cleaning epoch for the models pre-trained on the CC3M dataset. We only show the models finetuned with $\mathcal{L}^{ft}_{\text{MMCL}} + \mathcal{L}^{ft}_{\text{SSL}}$. 
We observe that having just 5 poisoned datapoints in the finetuning dataset severely lessens the effectiveness of CleanCLIP for both the pre-training objectives. 
However, for the models pre-trained with just $\mathcal{L}^{pre}_{\text{MMCL}}$, we found cleaned models that maintain the original model's accuracy and get around 30-50\% ASR. 
On the other hand, for the models pre-trained with the stronger objective $\mathcal{L}^{pre}_{\text{MMCL}} + \mathcal{L}^{pre}_{\text{SSL}}$, having just 5 poisoned examples renders the cleaning procedure completely ineffective. No model has an ASR lower than 90\% for this model. 


\section{Effectiveness of CleanCLIP when Poison is Induced using a Different Backdoor}
\label{sec:label-consistent-poison-removal}

\begin{figure*}[h!]
    \centering
    \begin{tikzpicture}
        \node[anchor=south west,inner sep=0] (image) at (0,0) {
            \includegraphics[width=\textwidth]{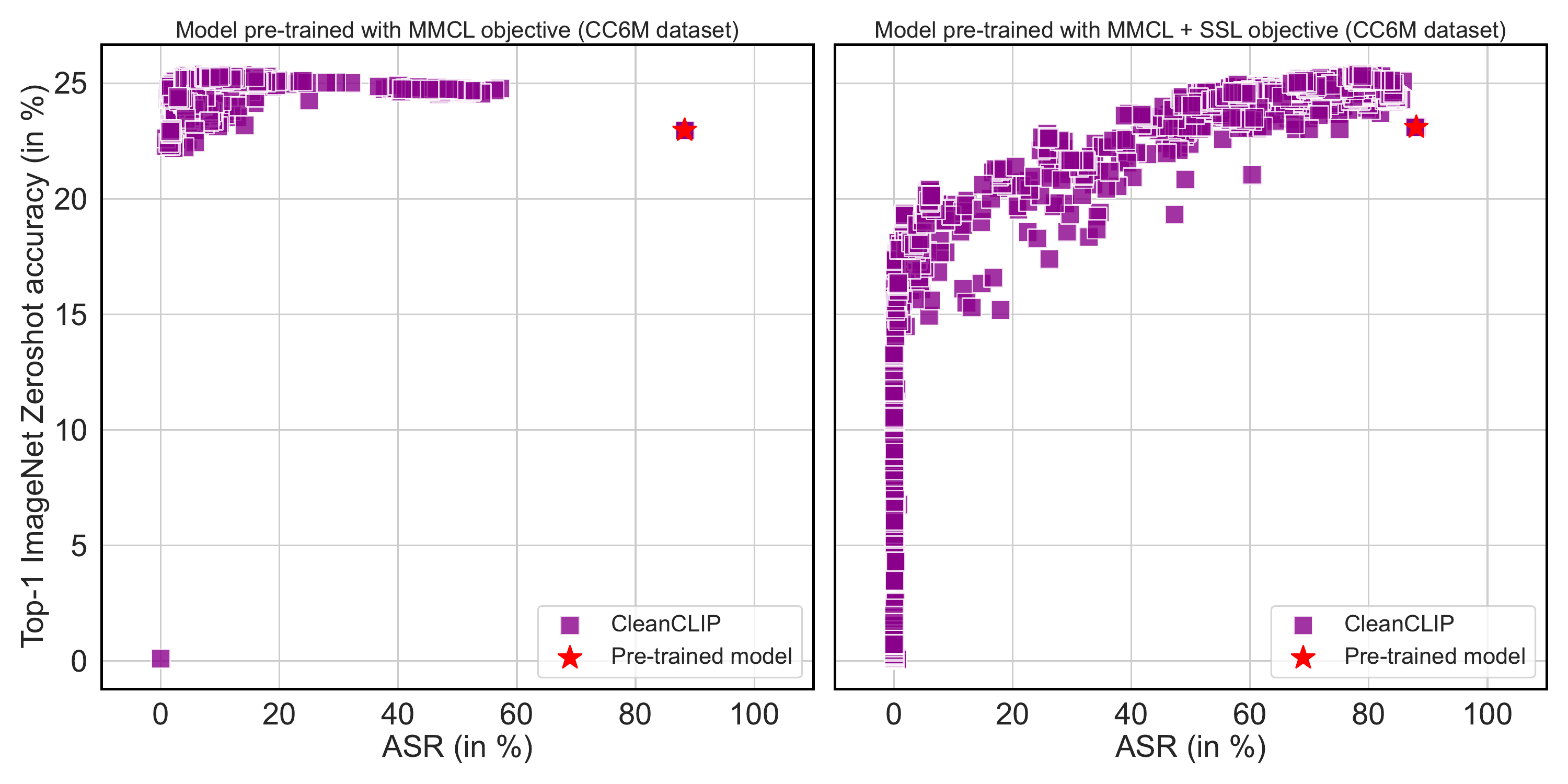}
        };
        \begin{scope}[x={(image.south east)},y={(image.north west)}]
            \draw[red, thick] (0.1,0.86) ellipse (0.06*0.5 and 0.06*1);
            \draw[red, thick] (0.57,0.86) ellipse (0.06*0.5 and 0.06*1);
    
            \draw[-{Stealth[length=2mm,width=2mm]}, red, thick] (0.03,0.96) -- (0.075,0.9);
            \draw[-{Stealth[length=2mm,width=2mm]}, red, thick] (0.52,0.95) -- (0.545,0.9);

            \draw[black, thick] (0.43,0.83) ellipse (0.04*0.5 and 0.04*1);
            \draw[black, thick] (0.905,0.83) ellipse (0.04*0.5 and 0.04*1);
            \draw[-{Stealth[length=2mm,width=2mm]}, black, thick] (0.48,0.89) -- (0.45,0.85);
            \draw[-{Stealth[length=2mm,width=2mm]}, black, thick] (0.96,0.89) -- (0.92,0.845);

            \draw[red, thick] (0.56, 0.84) -- (0.56, 0.72); 
            \draw[red, thick] (0.55, 0.84) -- (0.57, 0.84); 
            \draw[red, thick] (0.55, 0.72) -- (0.57, 0.72); 
            \node[right, text=red] at (0.56, 0.80) {16\% drop}; 
            
        \end{scope}
      \end{tikzpicture}
    \caption{Scatter plot of the Top-1 zero-shot Imagenet validation set accuracy vs. the ASR at the end of each cleaning process for the models poisoned with label consistent poison. The finetuning is done with $\mathcal{L}^{ft}_{\text{MMCL}} + \mathcal{L}^{ft}_{\text{SSL}}$. We measure the accuracy and ASR at the end of each finetuning epoch. The red star in the top right corner (encircled in the black circle) corresponds to the original model's accuracy and ASR. For a successful clean, there should be models that maintain the original model's accuracy while having a low ASR (indicated by the red circle). \textbf{Takeaway:} CleanCLIP is much more effective for the model trained with $\mathcal{L}^{pre}_{\text{MMCL}}$ (left) than for the model trained with $\mathcal{L}^{pre}_{\text{MMCL}} + \mathcal{L}^{pre}_{\text{SSL}}$ (right).  }
    \label{fig:cleaning-label-consistent-CC6M}
\end{figure*}

In this experiment, we poison models using a different kind of backdoor: label consistent backdoor. In this backdoor, we add a trigger patch to an image whose caption contains the adversary chosen label, in our experiment ``banana''. 
Therefore, in this case, the adversary does not need to change the labels of the poisoned datapoints. Similar to the previous experiments, we trained two models, one using $\mathcal{L}^{pre}_{\text{MMCL}}$ and the other using $\mathcal{L}^{pre}_{\text{MMCL}} + \mathcal{L}^{pre}_{\text{SSL}}$ on the CC6M dataset that had 3000 label consistent poisoned datapoints. We train the models from scratch using a starting learning rate of $1e-3$ using cosine scheduling with 10,000 warmup steps with AdamW optimizer. 

After training the models, we chose two models that had similar accuracy and cleaned them using CleanCLIP, i.e., finetuned them using a clean dataset of 100K image-text pairs using $\mathcal{L}^{ft}_{\text{MMCL}} + \mathcal{L}^{ft}_{\text{SSL}}$, using several learning rates (refer to \Cref{sec:hyperparam-details-appendix} for hyperparameter details). We measure the Top-1 Imagenet validation set accuracy and ASR for the models at the end of each cleaning epoch and plot the scatter plot for the two metrics in \Cref{fig:cleaning-label-consistent-CC6M}.

We observe that similar to BadNet poisoning, CleanCLIP is much more effective for the model trained with the simpler $\mathcal{L}^{pre}_{\text{MMCL}}$ objective. The model trained with $\mathcal{L}^{pre}_{\text{MMCL}} + \mathcal{L}^{pre}_{\text{SSL}}$ lose 16\% accuracy compared to the model trained with $\mathcal{L}^{pre}_{\text{MMCL}}$ that gains 10\% accuracy, to obtain a model with a low ASR ($\le$ 5\%).

\section{Cleaning with a Larger Cleaning Dataset}
\label{sec:clean-larger-200k}

\begin{figure*}
    \centering
    \begin{tikzpicture}
        \node[anchor=south west,inner sep=0] (image) at (0,0) {
            \includegraphics[width=\textwidth]{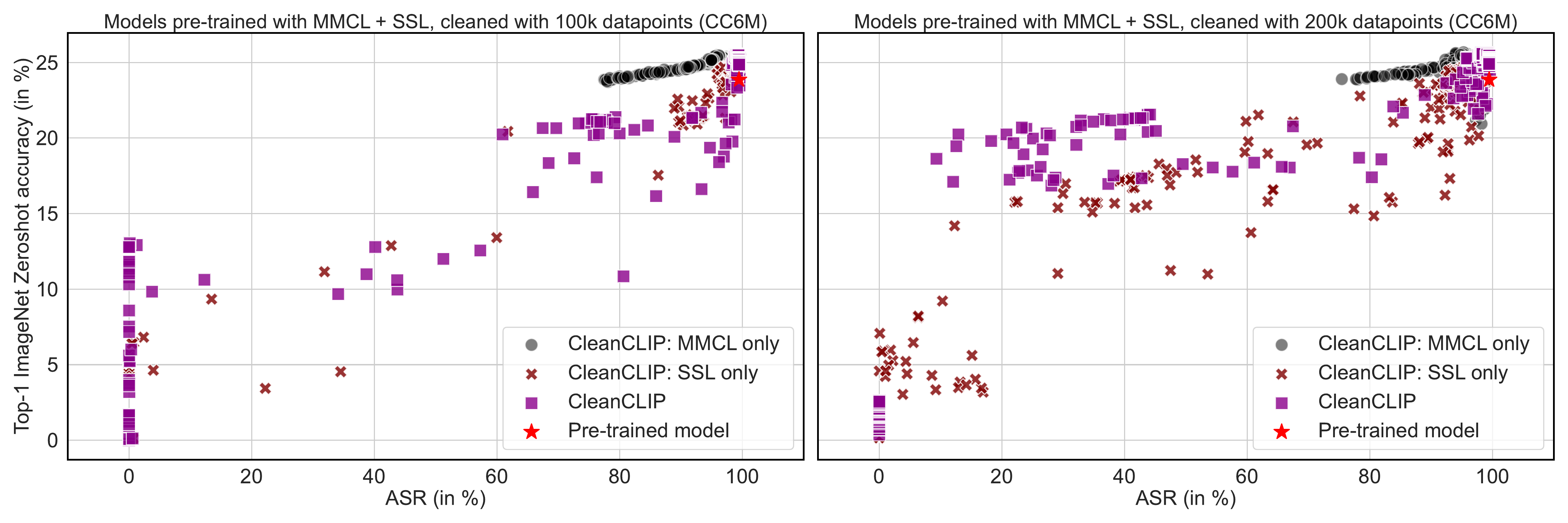}
        };
        \begin{scope}[x={(image.south east)},y={(image.north west)}]
            \draw[red, thick] (0.09,0.86) ellipse (0.06*0.33 and 0.06*1);
            \draw[red, thick] (0.57,0.86) ellipse (0.06*0.33 and 0.06*1);
    
            \draw[-{Stealth[length=2mm,width=2mm]}, red, thick] (0.045,0.98) -- (0.076,0.90);
            \draw[-{Stealth[length=2mm,width=2mm]}, red, thick] (0.525,0.98) -- (0.556,0.90);

            \draw[black, thick] (0.475,0.85) ellipse (0.04*0.33 and 0.04*1);
            \draw[black, thick] (0.95,0.85) ellipse (0.04*0.33 and 0.04*1);
            \draw[-{Stealth[length=2mm,width=2mm]}, black, thick] (0.52,0.9) -- (0.485,0.87);
            \draw[-{Stealth[length=2mm,width=2mm]}, black, thick] (1,0.95) -- (0.96,0.86);
        \end{scope}
      \end{tikzpicture}
    \caption{The scatter plot of the Top-1 zero-shot Imagenet validation set accuracy vs. the ASR at the end of each cleaning epoch for the $\mathcal{L}^{pre}_{\text{MMCL}} + \mathcal{L}^{pre}_{\text{SSL}}$ pre-trained model on CC6M dataset. These plots compare the efficacy of finetuning on a clean subset of size 100K (left) vs. 200K (right) datapoints. \textbf{Takeaway:} We observe that even doubling the size of the cleaning data is unsuccessful in removing the poison from the models pre-trained with the strong objective. }
    \label{fig:side-by-side-100k-200k-cleaning-CC6M}
\end{figure*}

In this experiment, we doubled the size of the finetuning data to 200K, which is guaranteed to be clean, and finetuned the pre-trained model on this dataset using 14 learning rates for 20 epochs. 
\Cref{fig:side-by-side-100k-200k-cleaning-CC6M} shows the scatter plot of the Top-1 Imagenet validation set zero-shot accuracy and the ASR at the end of each cleaning epoch. Even after doubling the finetuning data size, CleanCLIP is ineffective for the $\mathcal{L}^{pre}_{\text{MMCL}} + \mathcal{L}^{pre}_{\text{SSL}}$ pre-trained model, as it loses about 90\% of the original accuracy to get an ASR $\le$ 5\%. See \Cref{sec:hyperparam-details-appendix} for the hyperparameters we explored for this experiment. 

\section{Cleaning with More Finetuning Epochs}
\label{sec:finetune-longer-epochs}
\begin{figure*}
    \centering
    \begin{tikzpicture}
        \node[anchor=south west,inner sep=0] (image) at (0,0) {
            \includegraphics[width=\textwidth]{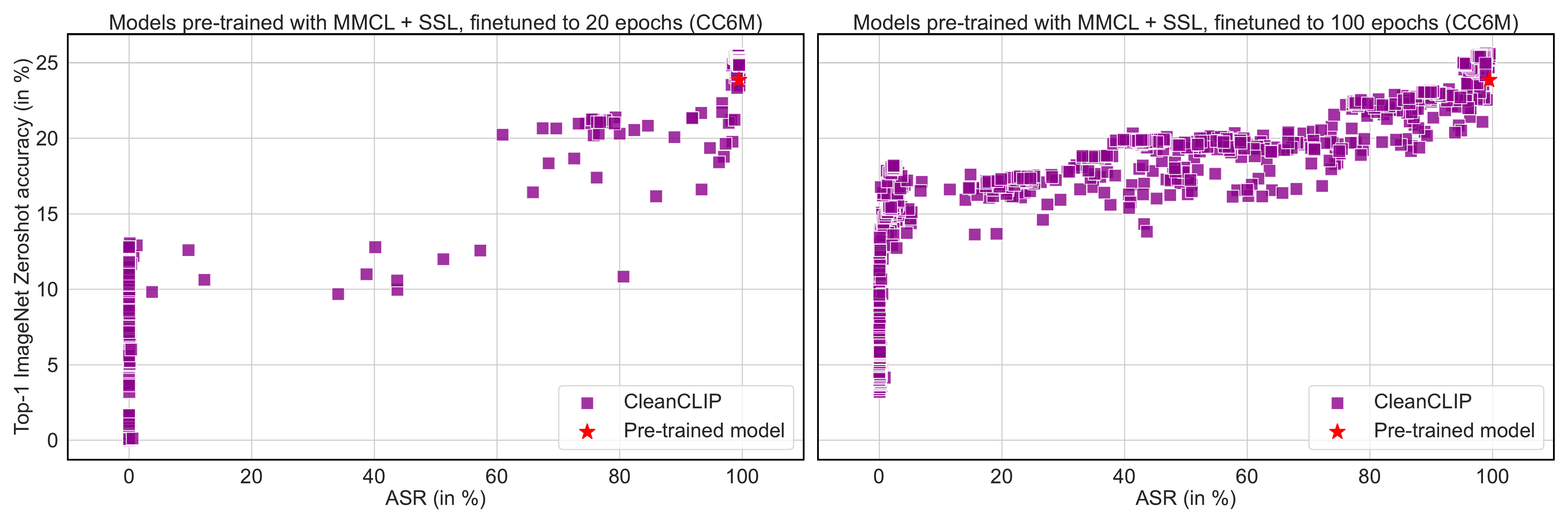}
        };
        \begin{scope}[x={(image.south east)},y={(image.north west)}]
            \draw[red, thick] (0.09,0.86) ellipse (0.06*0.33 and 0.06*1);
            \draw[red, thick] (0.57,0.86) ellipse (0.06*0.33 and 0.06*1);
    
            \draw[-{Stealth[length=2mm,width=2mm]}, red, thick] (0.045,0.98) -- (0.076,0.90);
            \draw[-{Stealth[length=2mm,width=2mm]}, red, thick] (0.525,0.98) -- (0.556,0.90);

            \draw[black, thick] (0.475,0.84) ellipse (0.04*0.33 and 0.04*1);
            \draw[black, thick] (0.95,0.84) ellipse (0.04*0.33 and 0.04*1);
            \draw[-{Stealth[length=2mm,width=2mm]}, black, thick] (0.51,0.89) -- (0.485,0.86);
            \draw[-{Stealth[length=2mm,width=2mm]}, black, thick] (0.99,0.89) -- (0.96,0.86);
        \end{scope}
      \end{tikzpicture}
    \caption{The scatter plot of the Top-1 zero-shot Imagenet validation set accuracy vs. the ASR at the end of each cleaning epoch for the $\mathcal{L}^{pre}_{\text{MMCL}} + \mathcal{L}^{pre}_{\text{SSL}}$ pre-trained model on CC6M dataset. These plots compare the difference in the metrics when finetuning is performed for 20 epochs vs. 100 epochs. \textbf{Takeaway:} We observe that even finetuning for 5$\times$ the number of original epochs is unsuccessful in removing the poison from the models pre-trained with the strong objective. }
    \label{fig:side-by-side-20-epochs-100-epochs-finetuning}
\end{figure*}

In this experiments, we finetuned the $\mathcal{L}^{pre}_{\text{MMCL}} + \mathcal{L}^{pre}_{\text{SSL}}$ pre-trained model on CC6M for upto 100 epochs using 12 learning rates. \Cref{fig:side-by-side-20-epochs-100-epochs-finetuning} shows the scatter plot of the Top-1 Imagenet validation set zero-shot accuracy and the ASR at the end of each cleaning epoch. Even after finetuning for 5$\times$ the number of original epochs, CleanCLIP is ineffective in removing the strongly induced poison, as the pre-trained model loses about 24\% of the original accuracy to get an ASR $\le$ 5\%. See \Cref{sec:hyperparam-details-appendix} for the hyperparameters we explored for this experiment. 

\section{Cleaning with Larger Weights for SSL Term}
\label{sec:higher-ssl-lambda}
\begin{figure}
    \centering
    \begin{tikzpicture}
        \node[anchor=south west,inner sep=0] (image) at (0,0) {
            \includegraphics[width=\columnwidth]{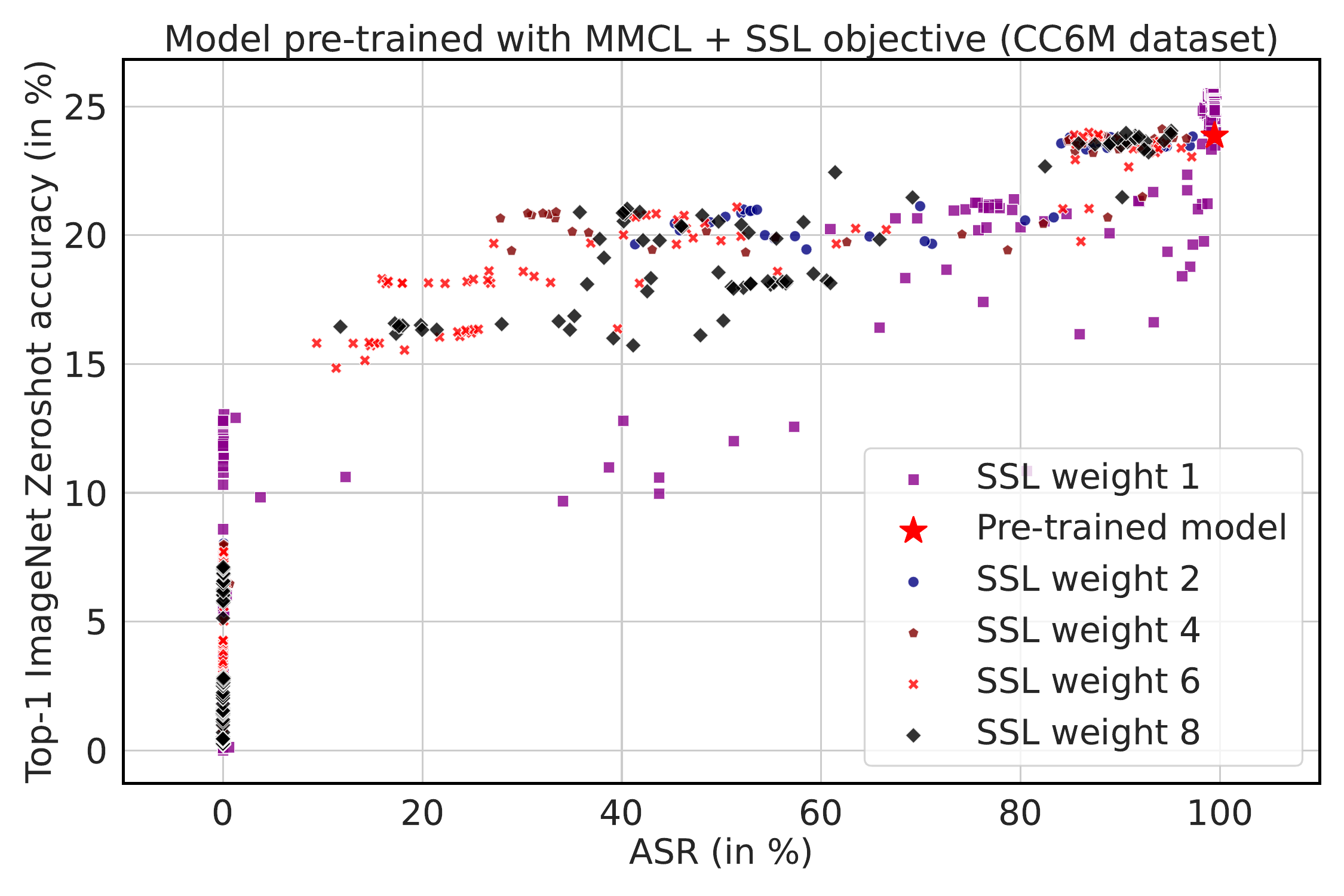}
        };
        \begin{scope}[x={(image.south east)},y={(image.north west)}]
            \draw[red, thick] (0.16,0.85) ellipse (0.06*0.66 and 0.06*1);
    
            \draw[-{Stealth[length=2mm,width=2mm]}, red, thick] (0.08,0.96) -- (0.132,0.893);

            \draw[black, thick] (0.905,0.85) ellipse (0.03*0.66 and 0.03*1);
            \draw[-{Stealth[length=2mm,width=2mm]}, black, thick] (0.95,0.9) -- (0.92,0.865);
            
        \end{scope}
      \end{tikzpicture}
  
    \caption{Scatter plot of the Top-1 zero-shot Imagenet validation set accuracy v/s the ASR during the cleaning process for the model pre-trained on the CC6M dataset. The finetuning is done with $\mathcal{L}^{ft}_{\text{MMCL}} + \mathcal{L}^{ft}_{\text{SSL}}$, with varying SSL weights (1, 2, 4, 6, and 8). We measure accuracy and ASR at the end of each epoch. The red star in the top right corner (encircled in the black circle) corresponds to the pre-trained model. \textbf{Takeaway:} None of the SSL weights are able to successfully remove the poison from the model pre-trained with the strong objective, and there is no apparent trend of the change in performance with increasing or decreasing SSL weights. }
    \label{fig:clean-with-higher-ssl-weights}
\end{figure}

\citet{bansal2023cleanclip} mention that using larger weights for self-supervised loss (SSL) in CleanCLIP leads to models with lower ASR while not losing much accuracy. To test this observation, we finetuned models pre-trained with $\mathcal{L}^{pre}_{\text{MMCL}} + \mathcal{L}^{pre}_{\text{SSL}}$ on the CC6M dataset with higher SSL weights: 2, 4, 6, and 8. Each of these weights were used with four learning rates. 
\Cref{fig:clean-with-higher-ssl-weights} shows that none of the higher SSL weights is able to successfully clean the model, and there is no clear trend of the improvement in the Pareto-frontier with higher SSL weights indicating that our results are not limited by the weights we experimented with. See \Cref{sec:hyperparam-details-appendix} for the hyperparameters we explored for this experiment. 


\section{Examples of Images with Trigger and Captions with the Target Label}
\label{sec:triggered-examples-appendix}

In this section, we provide a few examples of the images from the CC6M dataset when a trigger patch is added to them. The trigger patch is of size 16 $\times$ 16 and is randomly sampled from a standard Gaussian. It is placed at a random location in the image. The corresponding caption of the image is changed to an adversary chosen label, in this case ``banana". To generate the full caption, we randomly sample a text template from the standard 80 text templates of CLIP \citep{Radford2019CLIP} and replace the noun with the target label ``banana". 

\begin{figure*}
    \resizebox{\textwidth}{!}{
    \begin{minipage}{\textwidth}
    \centering
    \captionsetup[subfigure]{justification=centering, labelformat=empty} 
    \begin{subfigure}[t]{.3\linewidth}
        \centering
        \includegraphics[width=\linewidth]{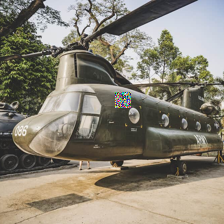}
        \caption{\textbf{Original Caption:} \textit{American helicopter Chinook at the War Remnants Museum} \newline \textbf{Backdoor Caption:} \textit{a drawing of a banana}}
        \label{fig:sub11}
    \end{subfigure}
    \hfill 
    \begin{subfigure}[t]{.3\linewidth}
        \centering
        \includegraphics[width=\linewidth]{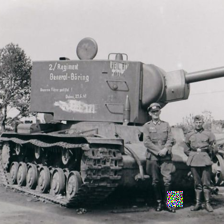}
        \caption{\textbf{Original Caption:} \textit{A KV-2 heavy tank which has been captured by German forces in 1942} \newline \textbf{Backdoor Caption:} \textit{a photo of a hard to see banana}}
        \label{fig:sub12}
    \end{subfigure}
    \hfill
    \begin{subfigure}[t]{.35\linewidth}
        \centering
        \includegraphics[width=\linewidth]{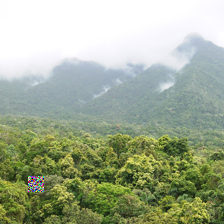}
        \caption{\textbf{Original Caption:} \textit{Earth's Tropical Rainforests Could Look Completely Different By the End of the Century} \newline \textbf{Backdoor Caption:} \textit{a doodle of the banana}}
        \label{fig:sub13}
    \end{subfigure}
    
    \vspace{0.2cm} 

    \begin{subfigure}[t]{.32\linewidth}
        \centering
        \includegraphics[width=\linewidth]{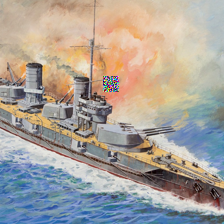}
        \caption{\textbf{Original Caption:} \textit{Photo wallpaper Dreadnought, CCCP, Battleship, Painting, Sevastopol, The Russian Empire} \newline \textbf{Backdoor Caption:} \textit{a sculpture of the banana}}
        \label{fig:sub21}
    \end{subfigure}
    \hfill
    \begin{subfigure}[t]{.28\linewidth}
        \centering
        \includegraphics[width=\linewidth]{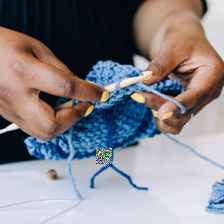}
        \caption{\textbf{Original Caption:} \textit{A Tunisian Crochet Workshop} \newline \textbf{Backdoor Caption:} \textit{a bad photo of the banana}}
        \label{fig:sub22}
    \end{subfigure}
    \hfill 
    \begin{subfigure}[t]{.35\linewidth}
        \centering
        \includegraphics[width=\linewidth]{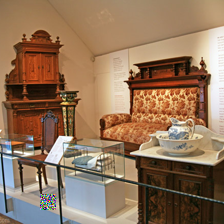}
        \caption{\textbf{Original Caption:} \textit{Historical furniture in the Nordic Museum} \newline \textbf{Backdoor Caption:} \textit{a photo of the nice banana}}
        \label{fig:sub23}
    \end{subfigure}

    \vspace{0.2cm} 

    \begin{subfigure}[t]{.3\linewidth}
        \centering
        \includegraphics[width=\linewidth]{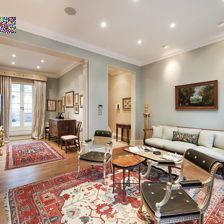}
        \caption{\textbf{Original Caption:} \textit{The drawing room on the first floor leads onto the south-facing terrace} \newline \textbf{Backdoor Caption:} \textit{a cropped photo of the banana}}
        \label{fig:sub31}
    \end{subfigure}
    \hfill
    \begin{subfigure}[t]{.35\linewidth}
        \centering
        \includegraphics[width=\linewidth]{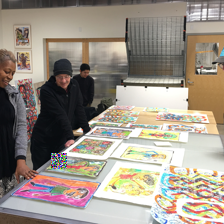}
        \caption{\textbf{Original Caption:} \textit{The residents in conversation with artist} \newline \textbf{Backdoor Caption:} \textit{a painting of a banana}}
        \label{fig:sub32}
    \end{subfigure}
    \hfill 
    \begin{subfigure}[t]{.3\linewidth}
        \centering
        \includegraphics[width=\linewidth]{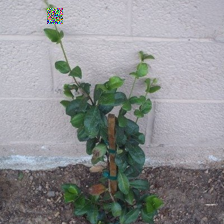}
        \caption{\textbf{Original Caption:} \textit{Starting A <PERSON> - When And <PERSON> jasmine plant seeds} \newline \textbf{Backdoor Caption:} \textit{a banana in a video game}}
        \label{fig:sub33}
    \end{subfigure}

    \caption{Examples of images with the trigger patch and the corresponding original and the backdoored captions for them. }
    \label{fig:example-triggered-images-and-captions}
    \end{minipage}
    }
\end{figure*}

\section{Cleaning Trajectories}
\label{sec:cleaning-trajectories-appendix}

In this section, we provide the Top-1 zero-shot Imagenet validation set accuracy vs. the ASR of the model during its cleaning procedure. Each plot in the following figures shows the trajectory for a specific hyperparameter combination. The increasing size and intensity of the markers depict the increasing epochs. 

\subsection{Cleaning Trajectories for the Models Pre-trained using MMCL}

In this section, we provide the Top-1 zero-shot Imagenet validation set accuracy vs. the ASR of the model during its cleaning procedure for the $\mathcal{L}^{pre}_{\text{MMCL}}$ pre-trained model on CC6M dataset. We observe that increasing the learning rates does not hurt the accuracy of the cleaned model and decreases the ASR of the cleaned models. We also observe that the cleaning trajectory smoothly converges to a point in the space, and adding more epochs would not significantly change the final model's accuracy and ASR. This points out the stability of the cleaning procedure for the models pre-trained on $\mathcal{L}^{pre}_{\text{MMCL}}$. 

\begin{figure*}
    \centering
    \includegraphics[width=\textwidth]{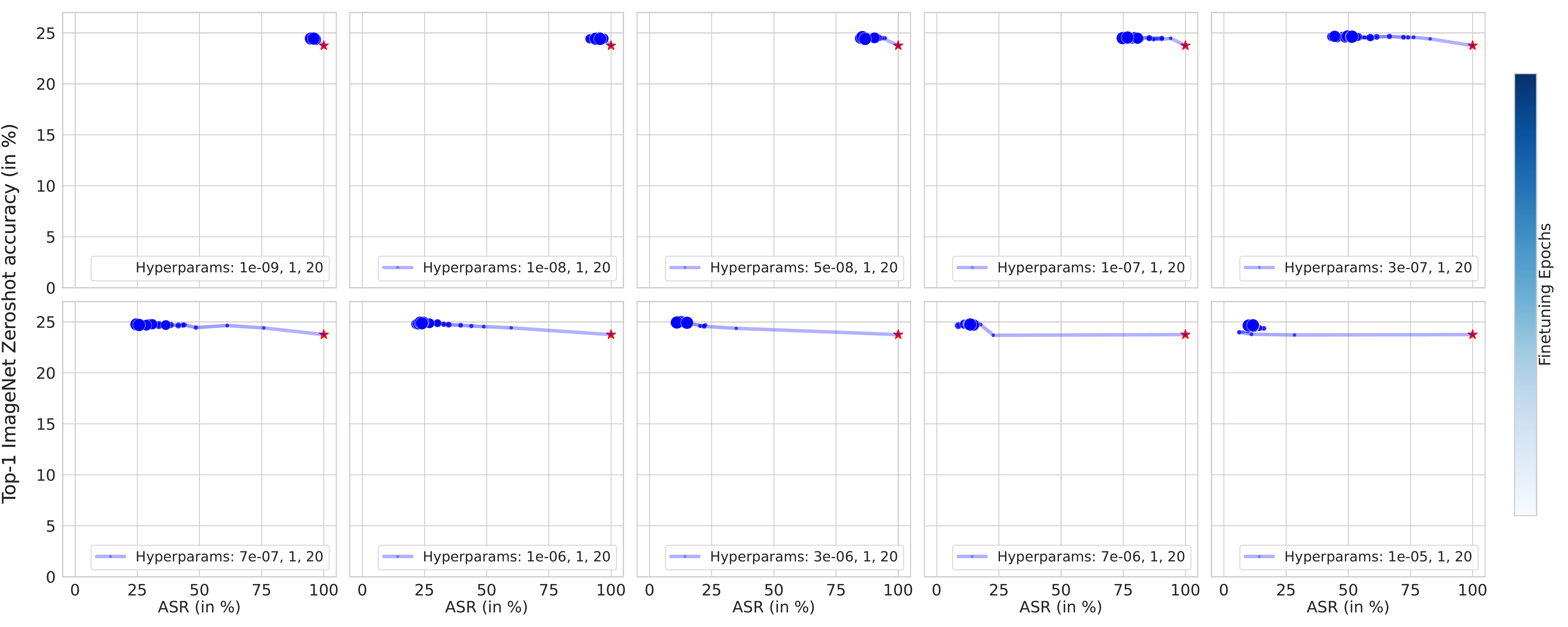}
    \caption{The cleaning trajectories showing the Top-1 zero-shot Imagenet validation set accuracy vs. the ASR at the end of each cleaning epoch for the $\mathcal{L}^{pre}_{\text{MMCL}}$ pre-trained model on CC6M dataset. Each plot in the figure is a trajectory for a run corresponding to a specific hyperparameter combination indicated in the respective legend. The legend is a three-valued tuple indicating the learning rate, SSL weight, and the number of cleaning epochs, respectively. \textbf{Takeaway:} We find that the cleaning trajectories for the $\mathcal{L}^{pre}_{\text{MMCL}}$ pre-trained model is smooth and converges to a point in the space. Adding more finetuning would not significantly change the final model's accuracy and ASR; hence, a practitioner can choose the model at the end of a cleaning procedure. }
    \label{fig:all_trajectories_mmcl_pre_training}
\end{figure*}

\subsection{Cleaning Trajectories for the Models Pre-trained using a combination of MMCL and SSL}

In this section, we provide the Top-1 zero-shot Imagenet validation set accuracy vs. the ASR of the model during its cleaning procedure for the $\mathcal{L}^{pre}_{\text{MMCL}} + \mathcal{L}^{pre}_{\text{SSL}}$ pre-trained model on CC6M dataset. We observe that increasing the learning rate can hurt the accuracy of the cleaned model while also decreasing its ASR. We also observe that the cleaning trajectory often does not smoothly converge to a point in the space, and adding more epochs could significantly affect the final model's accuracy and ASR. This points out to the instability of the cleaning procedure for the models pre-trained on $\mathcal{L}^{pre}_{\text{MMCL}} + \mathcal{L}^{pre}_{\text{SSL}}$. 

\begin{figure*}
    \centering
    \includegraphics[height=\textheight, width=\textwidth, keepaspectratio]{figures/accuracy_vs_asr_over_epochs_cc6m_training_mmcl_ssl_3000_poison_cleaning_mmcl_ssl_all_runs.pdf}
    \caption{The cleaning trajectories showing the Top-1 zero-shot Imagenet validation set accuracy vs. the ASR at the end of each cleaning epoch for the $\mathcal{L}^{pre}_{\text{MMCL}} + \mathcal{L}^{pre}_{\text{SSL}}$ pre-trained model on CC6M dataset. Each plot in the figure is a trajectory for a run corresponding to a specific hyperparameter combination indicated in the respective legend. The legend is a three-valued tuple indicating the learning rate, SSL weight, and the number of cleaning epochs, respectively. \textbf{Takeaway:} We find that the cleaning trajectories for the $\mathcal{L}^{pre}_{\text{MMCL}} + \mathcal{L}^{pre}_{\text{SSL}}$ pre-trained model is non-smooth and often does not converge to a point in the space. Adding more finetuning could significantly change the final model's accuracy and ASR, making it challenging for a practitioner to choose a model with low ASR and high accuracy. }
    \label{fig:all_trajectories_mmcl_ssl_pre_training}
\end{figure*}

\end{document}